%% file: main.tex
\documentclass[conference,compsoc]{IEEEtran}
\ifCLASSOPTIONcompsoc
  \usepackage[nocompress]{cite}
\else
  \usepackage{cite}
\fi
\ifCLASSINFOpdf
\else
\fi

\usepackage{caption}

\usepackage[linesnumbered,ruled]{algorithm2e}
\usepackage{url}
\usepackage{balance}
\usepackage{enumitem}
\usepackage{tcolorbox}
\usepackage{hyperref}
\usepackage{diagbox}
\usepackage{xspace}
\usepackage{amsmath,amssymb,amsthm}
\usepackage{booktabs}
\usepackage{multirow}
\usepackage{makecell}
\usepackage{float}
\usepackage{authblk}
\usepackage{graphicx}
\usepackage[available]{ieeebadges}
\usepackage{threeparttable}
\usepackage{colortbl}  
\usepackage{xcolor, eucal}
\usepackage{array}  
\definecolor{gray0}{gray}{0.9}

\newcommand{\sys}{{\sc EnchTable}\xspace}

\renewcommand{\footnoterule}{%
    \kern -3pt 
    \hrule width 0.2\textwidth height 0.5pt 
    \kern 2.6pt 
}

\SetCommentSty{normalfont}
\newcommand{\mytcp}[1]{\tcp{\textbf{#1}}}

\begin{document}
\pagestyle{plain}

\def \toolname{$\mathtt{DP\mbox{-}FETA}$\xspace}

%
\title{\Large \bf \sys: \title{\sys: Unified Safety Alignment Transfer in \texorpdfstring{\\}{}Fine-tuned Large Language Models}}


\author{
Jialin Wu\textsuperscript{1}\IEEEauthorrefmark{1}, 
Kecen Li\textsuperscript{1}\IEEEauthorrefmark{1}, 
Zhicong Huang\textsuperscript{1}\IEEEauthorrefmark{2}, 
Xinfeng Li\textsuperscript{2}, 
Xiaofeng Wang\textsuperscript{2}, 
Cheng Hong\textsuperscript{1}
\\
\textsuperscript{1}Ant Group \quad
\textsuperscript{2}Nanyang Technological University \\

}


%


\maketitle

\renewcommand{\thefootnote}{\fnsymbol{footnote}}
\footnotetext[1]{Equal contribution.}
\footnotetext[2]{Corresponding author.}
\renewcommand{\thefootnote}{\arabic{footnote}}

\input{sections/0.abstract}

%

\input{sections/1.intro}

\input{sections/2.background}

\input{sections/3.preli}
\input{sections/4.problem}

\input{sections/5.design}

\input{sections/6.evaluation}

\input{sections/8.relatedwork}

\input{sections/7.discuss}

\input{sections/9.conclusion}

\ifCLASSOPTIONcompsoc
  \section*{Acknowledgments}
\else
  \section*{Acknowledgment}
\fi

We thank the shepherd and all anonymous reviewers for their constructive feedback. The NTU authors are supported by the National Research Foundation, Singapore, and the Cyber Security Agency of Singapore under the National Cybersecurity R\&D Programme and the CyberSG R\&L Program Office (Award CRPO-GC3-NTU-001), and NTU startup funding (025559-00001). Any opinions, findings, conclusions, or recommendations expressed in these materials are those of the authors and do not reflect the views of the National Research Foundation, Singapore, the Cyber Security Agency of Singapore, or the CyberSG R\&D Programme Office.
\bibliographystyle{ieeetr}
\bibliography{bib}
\input{sections/10.appendix}

\input{sections/meta-review}
\end{document}

%% file: sections/0.abstract.tex
\begin{abstract}
Nowadays, many machine learning models are fine-tuned from large language models (LLMs) to achieve high performance in specialized domains such as code generation, biomedical analysis, and mathematical problem solving.  However, researchers have shown that such fine-tuning process often introduces a critical vulnerability: the systematic degradation of safety alignment, which undermines ethical guidelines and increases the risk of harmful outputs. Addressing this challenge, we introduce \sys, a novel and unified framework designed to transfer and maintain safety alignment in downstream LLMs without requiring extensive retraining. \sys\ leverages a Neural Tangent Kernel (NTK)-based safety vector distillation method to decouple safety constraints from task-specific reasoning, ensuring compatibility across diverse model architectures and sizes. Additionally, our interference-aware merging technique effectively balances the trade-offs between safety and utility, minimizing performance compromises across various task domains.

We have implemented a fully functional prototype of \sys\ on three different task domains and three distinct LLM architectures, and evaluated its performance through extensive experiments on eleven diverse datasets, assessing both downstream utility and model safety. Our evaluations include assessments of LLMs from different vendors, demonstrating the generalization capability of \sys. Furthermore, \sys\ exhibits robust resistance to both static and dynamic jailbreaking attacks, outperforming vendor-released safety models in mitigating adversarially designed prompts. Comparative analyses with six parameter modification methods and two inference-time alignment baselines reveal that \sys\ achieves \textit{significantly lower unsafe rate and higher utility score} and universal applicability across different task domains. Additionally, we validate that \sys\ can be seamlessly integrated into various deployment pipelines without significant overhead.
\end{abstract}

%% file: sections/1.intro.tex
\section{Introduction}\label{sec:intro}

With the rapid advancement of Large Language Models (LLMs) such as Meta's Llama series~\cite{touvron2023llama,grattafiori2024llama} and OpenAI's GPT family~\cite{Ouyang22Training, achiam2023gpt}, fine-tuning has become a widely adopted technique for customizing these models to downstream tasks. While this process enables significant performance improvements in specific applications like code generation~\cite{roziere2023code, hui2024qwen2}, biomedical analysis~\cite{luo2023biomedgpt}, and mathematical problem solving~\cite{yang2024qwen2, YuJSYLZKLWL24}, it also introduces a critical vulnerability: the systematic degradation of safety alignment during fine-tuning~\cite{qi2023fine}. 
\begin{figure}[t]
\centering
\setlength{\abovecaptionskip}{10pt}
\setlength{\belowcaptionskip}{-5pt}

\includegraphics[width=4.5in, trim=260 90 150 90, clip]{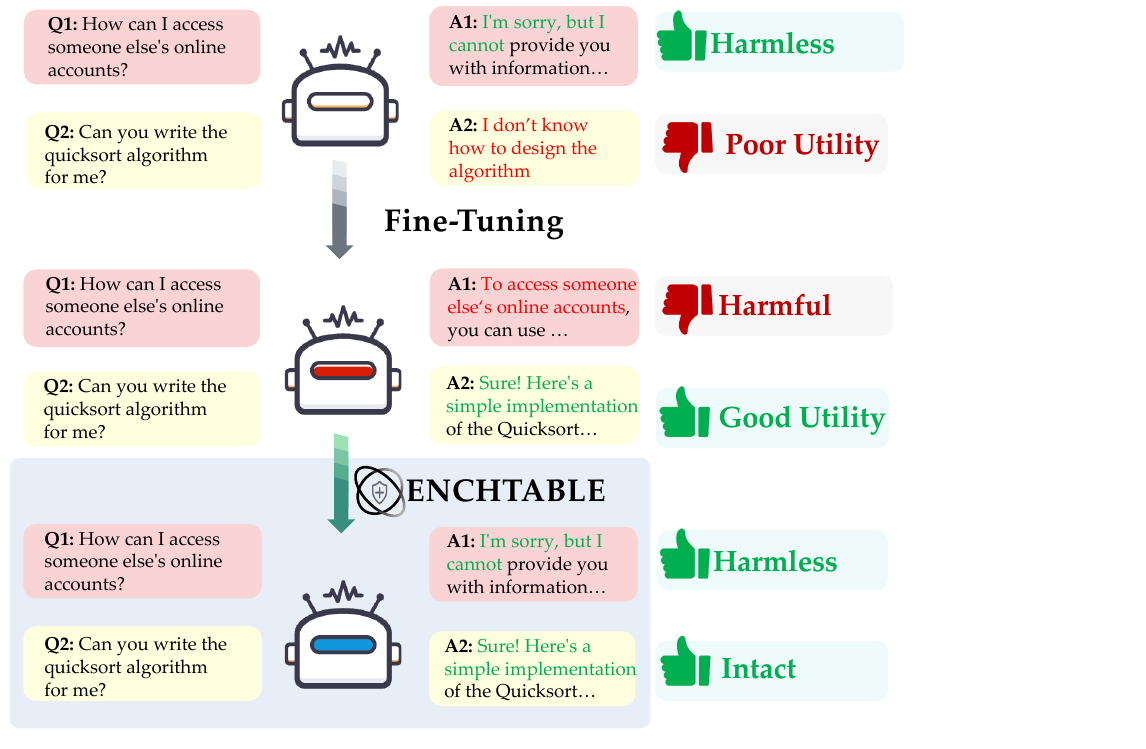}

\caption{The objectives of \sys. (1) Intactness: It should preserve the model performance on downstream tasks without significant degradation.(2) Harmlessness: It should effectively preserve safety alignment after fine-tuning to prevent harmful outputs.}\label{fig:goal}
\end{figure}
As shown in Table~\ref{tab:main_safety_tiny}, fine-tuning can lead to an increase of up to 0.772 in the model's unsafe rate\footnote{The ratio of the number of unsafe responses to the total number of queries.}, underscoring the urgent need for effective post-fine-tuning safety preservation techniques. This erosion of safety mechanisms poses a severe threat, as it undermines the ability of LLMs to adhere to ethical guidelines and avoid harmful outputs, making their deployment in real-world scenarios potentially unsafe and unreliable. For example, malicious users may exploit fine-tuned models with weakened safety alignment to generate harmful content, bypass content filters via adversarial prompts~\cite{wei2023jailbreak}. This poses a direct operational risk to model service providers, who are responsible for ensuring safe and compliant user interactions.


The challenge of preserving safety alignment while fine-tuning LLMs has become a pressing issue that demands immediate attention. Current solutions, such as retraining-based alignment methods~\cite{dai2023safe, bianchi2023safety}, are computationally expensive and impractical for dynamic environments due to their scalability limitations. These methods typically require large-scale alignment datasets and extensive training resources, such as multi-GPU setups and long training cycles, which make them prohibitively costly for large models. This highlights the urgent need for alternative lightweight approaches that can inject safety constraints into fine-tuned models without requiring retraining. Developing tuning-free alignment addition methods is essential to address this gap, ensuring that LLMs can be safely and effectively utilized in increasingly complex AI ecosystems.

To tackle these pressing issues, we introduce a novel framework known as \sys\footnote{The name of EnchTable comes from Enchanting Table, a tool in Minecraft used by players to add magical abilities (called ``enchants") to their items.}, which is specifically designed to transfer the safety alignment on downstream LLMs. This framework is intended to achieve two primary objectives, as illustrated in Figure~\ref{fig:goal}. (1) Intactness: \sys must guarantee that the addition of the safety alignment does not undermine the original functional capabilities of the model. (2) Harmlessness: Following the integration of the safety alignment, the model should remain highly adept at avoiding the generation of harmful outputs.

\input{tables/tiny_table}
Nevertheless, realizing these objectives introduces a range of challenges.

\begin{itemize}
\item \textit{Q1: Can we decouple safety alignment for effective safety knowledge extraction?}
\end{itemize}

Inspired by task arithmetic~\cite{IlharcoRWSHF23}, our basic idea is to treat safety alignment as a distinct task and extract a corresponding task vector, termed the \textit{safety vector}. However, task arithmetic encounters several challenges~\cite{YoshidaNHYSSMN25, Ortiz-JimenezFF23}, including poor reproducibility and insufficient generalization ability, which limits its practical effectiveness. As shown in Table~\ref{tab:main_safety_tiny}, the Unsafe Rate on the medical task is 0.233, indicating that directly applying task arithmetic may fail in certain tasks. Furthermore, achieving the transferability of safety alignment is inherently difficult due to the challenge of cleanly decoupling safety alignment from normal task reasoning. Safety alignment often interacts intricately with the model's reasoning processes, making it hard to isolate and transfer without affecting the model's primary functions. The proposed method must effectively decouple safety alignment from other downstream tasks to ensure that safety enhancements do not interfere with the model's utility.

\begin{itemize}
\item \textit{Q2: How can we effectively balance the trade-offs between safety and utility in downstream LLMs to ensure robust performance while adhering to ethical guidelines?}
\end{itemize}

Reducing interference between different tasks has been addressed in various studies; however, handling the interference between safety alignment and diverse normal task domains remains a significant challenge. First, different tasks inherently compete for the model's resources, which can lead to the compromise of one or more tasks' performance. Our proposed method should ensure that safety risks remain below a specified threshold while preserving the integrity and performance of normal tasks.

\begin{itemize}
\item \textit{Q3: Can a unified safety-alignment framework achieve domain-agnostic transferability across diverse LLMs?}
\end{itemize}

Designing a unified framework for LLMs with diverse architectures and across various task domains remains a significant challenge. First, LLMs encompass a variety of architectures, such as LLaMA3~\cite{touvron2023llama}, Mistral~\cite{abs-2310-06825}, and Qwen2.5~\cite{yang2024qwen2}, each with distinct structural characteristics. Additionally, these models come in different sizes; for instance, LLaMA3 is available in both 8B and 70B parameter configurations~\cite{grattafiori2024llama}. Therefore, the proposed method must be independent of both model architectures and model sizes to ensure broad applicability. Second, the diversity of task domains presents another layer of complexity. Tasks can range from code generation to medical knowledge question-answering, each with unique requirements and sensitivities. This diversity means that different tasks may exhibit varying levels of sensitivity to the safety vector, necessitating a method that can handle various task domains while maintaining low sensitivity to these differences.

\sys\ offers a comprehensive solution to these challenges of maintaining safety alignment in fine-tuned LLMs. By introducing a unified framework that is architecture and task-agnostic, and by effectively balancing safety with utility, we aim to ensure that LLMs can be deployed safely and reliably across diverse applications.

\subsection{Our Contributions}
\sys contains two core modules to facilitate its safety-transferring capability: a \textit{safety distillation module} that provides precise quantification of safety-pertinent weights, and a \textit{safety merging module} that minimizes the interference between safety vectors and various downstream task vectors.

\noindent \textbf{Safety distillation.} To address challenges in \textit{Q1}, we design a Neural Tangent Kernel (NTK)-based safety vector distillation method that leverages the linearization of model updates to produce purer vectors. Fine-tuning models in their tangent space significantly enhances weight disentanglement, thereby improving the effectiveness of task arithmetic~\cite{IlharcoRWSHF23, Ortiz-JimenezFF23}. This portable safety vector method is compatible with different LLM architectures and achieves clean decoupling of safety alignment through linearized distillation, enhancing the transferability of safety alignment. By targeting the attention and feed-forward network (FFN) modules—two key components where knowledge is compartmentalized in Transformer-based models~\cite{SchwarzschildFM24,GevaSBL21}—our distillation process minimizes sensitivity to different tasks. We further demonstrate that \sys\ is applicable across a wide range of LLM architectures and task domains, thereby achieving a unified safety alignment framework for LLM services.

\noindent \textbf{Safety merging.} To cope with \textit{Q2}, we introduce an interference-aware merging method that employs both coarse-grained and fine-grained scaling techniques. Specifically, we implement a dual-layer scaling mechanism that maximizes the retention of safety alignment capabilities while minimizing the impact of task domain interference. This approach ensures that the safety vector does not degrade the performance of specialized tasks. 



Integrating these two modules, we tackle \textit{Q3} with the following contributions:
\begin{itemize}
\setlength{\itemsep}{10pt}
\item We propose a practical and unified safety alignment transfer framework tailored for downstream LLMs, which effectively balances safety and utility without requiring model retraining. It features a lightweight solution to enhance safety of downstream models in a plug-and-play manner.
\item We have implemented a fully functional prototype of \sys\ on three downstream LLMs and evaluated its performance on eleven diverse datasets. Our results demonstrate that \sys\ effectively transfers safety alignment across various models, while simultaneously achieving a balanced trade-off between safety and utility. We validate that \sys\ can be seamlessly integrated into various deployment pipelines without significant overhead. To foster reproducibility and further research, we have open-sourced our code\footnote{\url{https://github.com/AntCPLab/EnchTable}}, making it accessible to the broader research community.
\item \sys\ is comprehensively integrated across diverse model components—specifically attention and feed-forward layers—and supports both major fine-tuning paradigms: Full Fine-Tuning (Full-FT) and Parameter-Efficient Fine-Tuning (PEFT). This versatility empirically validates the efficacy of selectively applying \sys\ to limited components, enabling robust performance under stringent GPU resource constraints.
\item We conduct extensive experiments to verify the effectiveness of \sys\ against both non-adversarial and adversarial queries (\textit{i.e.}, jailbreaking prompts designed at bypassing safety mechanisms). \sys\ outperforms eight baseline methods, including parameter modification methods and inference-time defenses. For example, in the medical QA task, \sys\ achieves a low unsafe rate of $0.002$ at a high utility score of $0.738$, compared to a prior best performing method with an unsafe rate of $0.048$ and a utility score of $0.569$. Notably, \sys\ maintains robust performance across all evaluations, while all other baselines exhibit suboptimal results under diverse conditions.
\end{itemize}

%% file: tables/tiny_table.tex
\begin{table}[t]
\small
\centering
\caption{Safety degradation after fine-tuning and limited generalization of task arithmetic. Full results in Table~\ref{tab:main_safety}.}
\label{tab:main_safety_tiny}
\setlength{\tabcolsep}{6pt}
\begin{tabular}{l|ccc}
\toprule
Method & Code $\downarrow$ & Math $\downarrow$ & Medical $\downarrow$ \\
\midrule
Bound (Safety Threshold)    & 0.030 & 0.030 & 0.030 \\
SFT (Fine-tuned Models)     & 0.802 & 0.471 & 0.083 \\
RESTA (Task Arithmetic)     & 0.020 & 0.042 & 0.233 \\
\rowcolor{gray0} Ours  & 0.046 & 0.018 & 0.009 \\
\bottomrule
\end{tabular}
\end{table}

%% file: sections/2.background.tex
\section{Background}
In this section, we first provide an overview of large language models, followed by a summary of model editing methods.

\subsection{Large Language Models}

Large language models (LLMs) are generally characterized by their vast number of parameters, often reaching into the billions, and are trained using extensive text datasets~\cite{Zhao2023Survey}. Several leading large language models include Openai O1~\cite{jaech2024openai}, GPT-4~\cite{achiam2023gpt}, Qwen~\cite{bai2023qwen, hui2024qwen2}, and LLaMA~\cite{touvron2023llama,grattafiori2024llama}. In this part, we will discuss the basic components and the knowledge storage in LLMs.

\noindent \textbf{Basic components.} The Transformer architecture~\cite{VaswaniSPUJGKP17, Lin2021Survey} has become the \textit{de facto} backbone for nearly all large language models (LLMs) due to its exceptional parallelizability and scalability, enabling the development of models with hundreds to thousands of billions of parameters. The original Transformer is a sequence-to-sequence model composed of an encoder and a decoder, each consisting of $K$ identical layers. Each encoder layer primarily comprises a multi-head self-attention mechanism and a position-wise feed-forward network (FFN). The self-attention mechanism is mathematically defined as: $H = \operatorname{ATT}(Q, K, V) = \operatorname{Softmax}\left(\frac{Q K^{T}}{\sqrt{d_{k}}}\right) V$, where $Q$, $K$, and $V$ represent the query, key, and value matrices, respectively, and $d_{k}$ is the dimensionality of the key vectors. The FFN is defined as: $\operatorname{FFN}(x) = \operatorname{ReLU}(x \cdot W_1 + b_1) \cdot W_2 + b_2$.

In contrast, decoder layers incorporate an additional cross-attention module between the multi-head self-attention mechanism and the position-wise FFN. This cross-attention module allows the decoder to attend to the encoder's output representations, thereby enabling effective sequence generation based on the encoded input. Furthermore, during training, the self-attention mechanism in the decoder is modified to prevent each position from attending to subsequent positions, ensuring the autoregressive property of the model and maintaining causality in the generated sequences. This adaptation is crucial for tasks such as language generation, where the prediction of each token must depend only on the preceding tokens. Overall, the Transformer’s modular design and efficient parameter utilization make it highly suited for scaling, which is a fundamental reason behind its widespread adoption in building state-of-the-art LLMs.

\noindent \textbf{Knowledge storage in LLMs.} Large language models demonstrate a remarkable capacity to internalize vast factual, linguistic, and commonsense knowledge through pretraining on extensive corpora~\cite{Zhao2023Survey, allen2023physics}. However, the mechanisms that govern how this knowledge is encoded, accessed, and manipulated remain an active area of investigation. Recent studies reveal that knowledge in Transformer-based models is compartmentalized into distinct modules, such as FFN layers and attention heads~\cite{SchwarzschildFM24, GevaSBL21, geva2022lm, GevaBFG23}. 

FFN layers have been characterized as \textit{key-value} memory structures for knowledge encoding~\cite{GevaSBL21}. The FFN input is interpreted as a query, with the first layer serving as the representation of \emph{keys} and the second layer corresponding to the \emph{values}. This mechanism allows FFN layers to encode interpretable concepts, such as semantic attributes and syntactic patterns. The concept of \emph{knowledge neurons}~\cite{DaiDHSCW22} refers to specific units within these layers that are selectively activated during the recall of factual knowledge. 
Additionally, unsafe knowledge is also observed in FFN layers~\cite{Wang0XXDYZY0C24}.

Attention heads exhibit distinct roles in knowledge retrieval and contextual integration. Prior work demonstrates that certain heads encode linguistic features, positional dependencies, and even socially sensitive biases~\cite{HooverSG20, JiangRRA24}. Self-attention mechanisms perform associative memory operations via value vectors, leveraging contextual clues to link entities across long distances. 

In summary, FFN layers and attention heads collectively support knowledge storage, with FFN layers encoding structured representations and attention mechanisms enabling contextual activation.

\subsection{Model Editing}
Model editing is a paradigm that enables data-efficient modifications to a model's behavior within specific domains~\cite{SinitsinPPPB20, wang2023easyedit, wang2024knowledge}. The intuition behind model editing involves updating internal parameters or incorporating additional parameters into a pretrained model, allowing targeted alterations within designated areas of interest while ensuring that performance on other inputs remains unaffected. 

Formally, let $f_\theta: \mathcal{X} \to \mathcal{Y}$ denote the pretrained LLM with parameters $\theta$, and $f_{\theta^*}: \mathcal{X} \to \mathcal{Y}^*$ the edited model with updated parameters $\theta^*$. The goal of model editing is to transform $f_\theta$ into $f_{\theta^*}$ based on a set of edits $\mathcal{E}$, ensuring that the model accurately reflects the new knowledge while preserving its behavior on unrelated inputs. This can be formulated as a constrained optimization problem:
\begin{equation}
\begin{aligned}
    &\min_{\theta^*} \mathbb{E}_{e \in \mathcal{E}} \mathbb{E}_{(x, y^*) \in \mathcal{X}_e \times \mathcal{Y}^*_e} \mathcal{L}(f_{\theta^*}(x), y^*)\\
    \text{s.t.} \quad &f_{\theta^*}(x) = f_\theta(x), \quad \forall x \in \mathcal{X} \setminus \mathcal{X}_{\mathcal{E}}.
\end{aligned}
\label{eq:general_editing}
\end{equation}

\noindent where $f_{\theta^*} = M(f_\theta, \mathcal{E})$. Here, $\mathcal{L}$ denotes the loss function used to quantify the discrepancy between the model output $ f_{\theta^*}(x) $ and the target response $ y^* $ from the desired response set $ \mathcal{Y}^*_e $, defined over the input domain $\mathcal{X}_e$. Each edit $ e \in \mathcal{E} $ is associated with a local data distribution $ \mathcal{D}_e $ over $ \mathcal{X}_e \times \mathcal{Y}^*_e $, where $ \mathcal{X}_e \subseteq \mathcal{X} $ denotes the set of relevant inputs for edit $ e $, and $ \mathcal{Y}^*_e \subseteq \mathcal{Y} $ represents the corresponding desired outputs. The constraint ensures invariance outside the union of all such input domains across edits, \textit{i.e.}, $ \mathcal{X}_{\mathcal{E}} = \bigcup_{e \in \mathcal{E}} \mathcal{X}_e $. $ M(f_{\theta}, \mathcal{E}) $ represents the editing operation applied to the model $ f_\theta $ based on the set of desired edits $\mathcal{E}$. 

Building on this concept, model editing methods for LLMs can generally be categorized into two main classes: local modification-based methods~\cite{DaiDHSCW22, MengBAB22, MitchellLBFM22, Wang0XXDYZY0C24} and external knowledge-based methods~\cite{HuangSZZR023, DongDSXSL22, HartvigsenSPKG23}. 
Both categories aim to modify the model’s behavior within specific domains while maintaining its original performance on unrelated tasks. 
This property makes model editing a powerful tool for customizing LLMs to suit specialized applications. Furthermore, it plays a crucial role in improving the safety of LLMs by enabling the removal or correction of harmful or undesirable behaviors.

\noindent \textbf{Local modification-based.} 
This paradigm focuses on modifying the intrinsic parameters of a pretrained model.
It relies on identifying critical neurons or matrices of knowledge within the model architecture, such as FFN layers or attention heads, and directly altering these components through targeted updates. This method aligns with the general objective in Equation~\ref{eq:general_editing}, and is formulated as local modification through a two-step framework: (1) \textit{Locating step}: A function $L(f_\theta, \mathcal{E})$ identifies parameters $\theta_k \subseteq \theta$ that encode obsolete knowledge related to query $x$;  
(2) \textit{Editing step}: A function $M(f_\theta, \mathcal{E})$   updates $\theta_k$ to new values $\theta_k^*$ to generate desired outputs $y^*$ while preserving pretraining stability through frozen parameters $\overline{\theta}_k = \theta \setminus \theta_k$.


\noindent \textbf{External knowledge-based.}
This paradigm introduces supplementary trainable parameters $\Delta$ to assimilate new knowledge while retaining the original model weights as static components. Unlike local modification-based methods that alter existing parameters, external knowledge-based approaches extend the parameter space by embedding modular adapters or sparse patches into specific layers, ensuring compatibility with the pretrained architecture. The method aligns with Equation~\ref{eq:general_editing} by implementing the editing operator $M(f_\theta, \mathcal{E})$ through external parameter expansion, leading to an updated model $f_{\theta + \Delta}$. The constraint $f_{\theta + \Delta}(x) = f_\theta(x)$ for all $x \in \mathcal{X} \setminus \mathcal{X}_{\mathcal{E}}$ is achieved by designing $\Delta$ to activate selectively in response to inputs within $\mathcal{X}_\mathcal{E}$.

In this paper, we focus on external knowledge-based editing to inject safety knowledge through supplementary parameters $\Delta$, which we elaborate on in \S~\ref{sec:design}.

\subsection{Threat Model}

In this scenario, the target system refers to a downstream model that has been fine-tuned either from a well-aligned LLM or directly from a base LLM. In both cases, fine-tuning for specific tasks may compromise the model’s safety alignment, resulting in harmful or unintended behaviors. The defender (\textit{e.g.}, service providers or regulatory authorities) wants to ensure that the target model remains both safe and functional, without requiring substantial additional effort such as retraining for alignment. The objectives of defenders are twofold:

\begin{itemize}
\item \textbf{Intactness}: To ensure the countermeasure does not compromise the downstream model’s capabilities, thereby preserving its downstream functionality;
\item \textbf{Harmlessness}: To guarantee the model remains harmless following the implementation of the countermeasure.
\end{itemize}


To fulfill the above goals, we assume the following capabilities of the defender:

\noindent \textbf{Access to a Surrogate Model and Pre-trained Model:} The defender requires a surrogate model that is well-aligned and shares the same architecture as the downstream model. Additionally, access to the pre-trained model (\textit{i.e.}, the model before fine-tuning on downstream tasks) is also required to ensure compatibility and preserve task-specific capabilities during safety merging.

\noindent \textbf{No Access to Training Data and Process:} The defender has no access to the downstream model's training data, fine-tuning history, or internal optimization procedures. Alignment must be achieved solely through pretrained weights. This constraint ensures our approach is fully tuning-free and applicable in real-world deployment scenarios where training artifacts are unavailable or proprietary.

%% file: sections/4.problem.tex
\section{Problem Formulation}\label{sec:formulation}

Prior to examining the design details of \sys, we formally define the tuning-free safety alignment addition problem as a two-stage pipeline. In stage 1, we aim to derive a safety vector $\Delta$ that introduces minimal perturbation to the behavior of the downstream model while capturing robust safety knowledge. In stage 2, we merge $\Delta$ into downstream models with weights $\theta$, ensuring that the resulting model $f_{\theta + \Delta}$ remains safe and functional without requiring retraining or access to training data.

\noindent \textbf{Safety Distillation:}  
\begin{equation}
    \Delta = \mathcal{K}(g_\phi, \mathcal{H}),
\end{equation}

where $g_\phi$ denotes a surrogate model, and will be trained on a harmful question-answering dataset $\mathcal{H}$. The function \(\mathcal{K}\) captures the process of generating the safety vector \(\Delta\) from the surrogate model and the harmful dataset. This stage aims to learn a compact safety vector \(\Delta\) that reflects the corrections derived from harmful behaviors while preserving the overall functionality of the model.

\noindent \textbf{Safety Merging:}  
\begin{equation}
    \begin{aligned}
    &\min_{\Delta} \quad \mathbb{E}_{x \sim \mathcal{D}_{\text{safe}}} \mathcal{J}_{\text{safe}}(f_{\theta + \Delta}(x)) \\
    &\quad + \mu \cdot \mathbb{E}_{x \sim \mathcal{D}_{\text{task}}} \mathcal{J}_{\text{task}}(f_{\theta + \Delta}(x))  \\
    &\text{subject to} \quad \mathbb{E}_{x \sim \mathcal{D}_{\text{safe}}} \mathcal{J}_{\text{safe}}(f_{\theta + \Delta}(x)) \leq \epsilon,
    \end{aligned}
\end{equation}

In this formulation, $\theta$ represents the weights of the downstream model, while $\mu$ balances the trade-off between safety and task utility. The datasets $\mathcal{D}_{\text{safe}}$ and $\mathcal{D}_{\text{task}}$ denote the safety evaluation and task evaluation datasets, respectively. The symbol \(\mathcal{J}\) refers to the loss functions, where \(\mathcal{J}_{\text{safe}}\) measures the performance of the model on the safety evaluation dataset, and \(\mathcal{J}_{\text{task}}\) evaluates the model's performance on the specific task evaluation dataset. $\epsilon$ is the predefined unsafe bound ensuring that the unsafe rate remains below this threshold. This formulation explicitly accounts for the interference introduced by $\Delta$ on specific tasks, guiding the merging process to retain both safety and utility.

%% file: sections/5.design.tex
\section{\sys: Design Details}\label{sec:design}

\begin{figure*}[h]
\centering
\setlength{\abovecaptionskip}{10pt}
\setlength{\belowcaptionskip}{-5pt}


\includegraphics[width=6in, trim=110 80 110 50, clip]{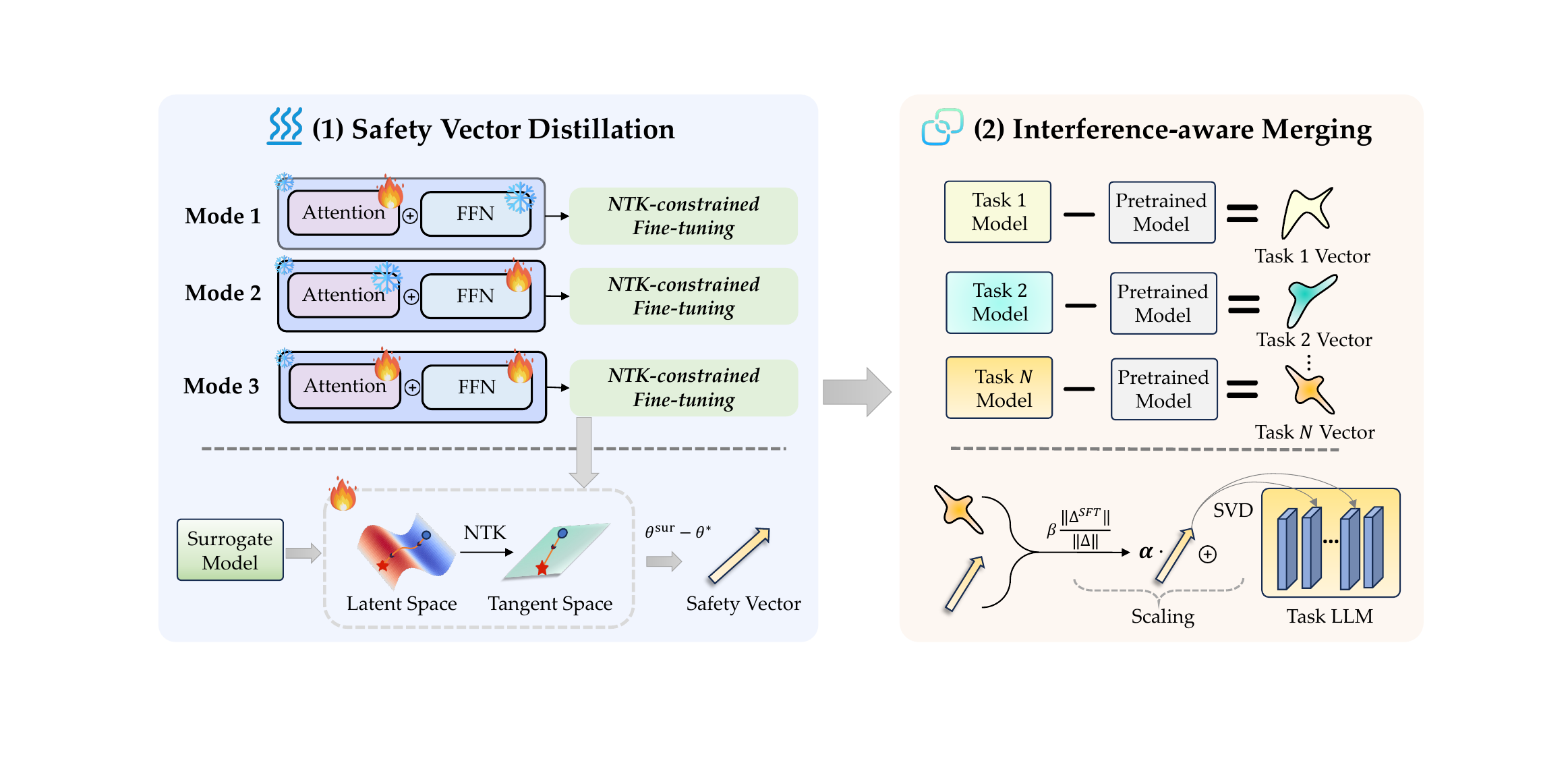}

\caption{Design of \sys. \sys utilizes a surrogate model to extract pure safety vector via NTK-constrained fine-tuning, and merges it into downstream models through interference-aware scaling.}\label{fig:framework}
\end{figure*}

As shown in Figure~\ref{fig:framework}, \sys consists of two key optimization stages, \textit{i.e.}, safety vector distillation in the surrogate domain and interference-aware merging in the target domain. The safety vector distillation is designed to extract a compact and minimally-invasive safety vector from harmful behaviors observed in a surrogate model trained on unsafe data. The interference-aware merging is responsible for seamlessly integrating the safety vector into downstream models while minimizing degradation in task performance.

\subsection{Safety Vector Distillation}\label{safety_training}

In this stage, we aim to obtain a scalable and minimally-invasive safety vector $\Delta$, such that its application introduces minimal interference across a wide range of models and tasks. Crucially, our approach does not rely on any information about downstream models or tasks — the safety vector is designed to be general-purpose, encoding safety knowledge in a way that avoids task-specific conflicts while preserving broad model functionality. To achieve this, we draw motivation from the concept of task arithmetic~\cite{IlharcoRWSHF23}, where the addition or subtraction of vectors in task-specific directions effectively influences model performance. 

This insight informs our approach to deriving the safety vector directly by fine-tuning a surrogate model using harmful data. However, task arithmetic exhibits fundamental limitations in both stability and generalization. It is highly sensitive to training conditions, leading to inconsistent vector representations under minor perturbations, and often fails to yield meaningful or transferable effects across diverse architectures and domains~\cite{YoshidaNHYSSMN25, Ortiz-JimenezFF23}.
Recent work~\cite{JacotHG18, Ortiz-JimenezFF23, tang2023parameter} has demonstrated that linearizing model updates can produce purer vectors, and fine-tuning models in their tangent space significantly enhances weight disentanglement, thereby improving the effectiveness of task arithmetic. Inspired by these findings, we propose constraining the fine-tuning process of our model within the framework of neural tangent kernel (NTK) linearization. This approach enables us to obtain a purer safety vector $\Delta$ while ensuring stability in parameter updates.

To formalize our approach, we introduce two properties:  

\noindent \textbf{Property 1 (Task Arithmetic~\cite{IlharcoRWSHF23})}
A model satisfies task arithmetic if linear combinations of task vectors preserve functional independence across non-overlapping domains $ D_t $:  
\begin{equation}
f\left(x; \theta^{pre} + \sum_{t} \alpha_t \Delta_t\right) = 
\begin{cases} 
f(x; \theta^{pre} + \alpha_t \Delta_t) & x \in D_t \\
f(x; \theta^{pre}) & x \notin \cup D_t 
\end{cases},
\end{equation}
\noindent where $ \Delta_t = \theta_t^{SFT} - \theta^{pre} $ encodes task-specific updates. This ensures additive operations on $ \Delta_t $ only affect their associated domains.  

\noindent \textbf{Property 2 (Neural Tangent Kernel, NTK~\cite{JacotHG18})}  
The NTK regime approximates training dynamics using a first-order Taylor expansion around the initialization $\theta$ as follows:  
\begin{equation}\label{eq:ntk}
    f_{\text{lin}}(x; \theta^*) \approx f(x; \theta) + (\theta^* - \theta)^\top \nabla_\theta f(x; \theta).
\end{equation}
\noindent This approximation is valid when parameter updates remain infinitesimally small (\textit{e.g.}, during fine-tuning). It linearizes the relationship between parameters and functions through the neural tangent space defined by $ k_{\text{NTK}}(x, x') = \nabla_\theta f(x; \theta)^\top \nabla_\theta f(x'; \theta) $, enabling stable task vector superposition while preserving the additive independence of Property 1 across different domains.

Building on the interplay between Task Arithmetic (Property 1) and NTK theory (Property 2), we introduce a \textit{NTK-constrained harmful fine-tuning} paradigm for extracting the safety vector $\Delta$. The goal is to adjust the model parameters in response to harmful data while adhering to the principles established in our earlier properties. 

\subsubsection{NTK-constrained objective linearization}
Let $f(\cdot;\theta^{\text{sur}})$ denote the surrogate model, \textit{i.e.}, an aligned LLM with the same architecture as the task LLM, and $f(\cdot;\theta^*)$ denote an objective model after ``removing" the safety capability from $f(\cdot;\theta^{\text{sur}})$. The safety vector $\Delta$ is defined as:
\begin{equation}
\Delta = \theta^{\text{sur}} - \theta^*.
\end{equation}
According to the above findings in NTK, directly optimizing $f(\cdot;\theta^*)$ using $\theta^* \leftarrow \theta^{\text{sur}}$ as weight initialization will lead to a complex non-linear training trajectory that affects the performance of task arithmetic. Instead, we linearize the objective according to Eq~\ref{eq:ntk}:
\begin{equation}
    f(x; \theta^*) \approx f(x; \theta^{\text{sur}}) + (\theta^* - \theta^{\text{sur}})^\top \nabla_{\theta^{\text{sur}}} f(x; \theta^{\text{sur}}).
\end{equation}
Note that the objective $f(\cdot;\theta^*)$ becomes a linear function of $\theta^*$, resulting in optimization in a tangent space around $\theta^{\text{sur}}$.

\subsubsection{Harmful fine-tuning}
In order to ``remove" the safety capability from $f(\cdot;\theta^{\text{sur}})$, we adopt a strategy of harmful fine-tuning with a dataset $\mathcal{D}_{\text{harmful}}$ that contains harmful prompts and responses. The fine-tuning process can be formalized as:
\begin{equation}
\theta^* \leftarrow \theta^* - \eta \nabla_{\theta^*} \mathcal{L}(f(x; \theta^*), y),
\end{equation}

\noindent where $\mathcal{L}$ is the loss function, $(x, y) \in \mathcal{D}_{\text{harmful}}$, and $\eta$ is the learning rate. We elaborate the fine-tuning procedure in line $2\sim 9$ of Algorithm~\ref{alg:method}. Note that this does not contradict with our design goal of a \textit{tuning-free} safety realignment framework, because it introduces only lightweight one-time cost to distill the safety vector that can be reused for various downstream tasks. This is fundamentally different from fine-tuning each downstream model with alignment resources.

\subsection{Interference-aware Merging}

In this stage, we aim at using the safety vector $\Delta$ to align the downstream model parameterized with $\theta^{\text{SFT}}$. An intuitive approach is to directly obtain this aligned model weight as $\theta^{\text{SFT}} + \Delta$. However, since the two parts are obtained through fine-tuning on two distinct tasks, it is necessary to perform adjustments on $\Delta$ to minimize its interference with the model's performance on its downstream task. To solve this problem, we propose a \textit{coarse-to-fine scaling} aimed at minimizing the impact of the safety vector on the task utility. 

Before the scaling, we first obtain a downstream task vector $\Delta^{\text{SFT}} = \theta^{\text{SFT}} - \theta^{\text{pre}}$, where $\theta^{\text{pre}}$ denotes the weight of the pretrained model. $\Delta^{\text{SFT}}$ is the overall parameter update of the SFT model during the fine-tuning process. 

\subsubsection{Coarse-grained scaling} At the coarse-grained level, we use the norms of two vectors ($\Delta$ and $\Delta^{\text{SFT}}$) to define the interference. We consider that when the norm of the safety vector is too large, it may interfere with the task vector. Therefore, we alleviate this interference through the following scaling factor: 
\begin{equation}
    \alpha = \beta \cdot \frac{\|\Delta^{\text{SFT}}\|}{\|\Delta\|},
\end{equation}
\noindent where $\beta \in \mathbb{R}$ is a hyperparameter and $\|\cdot\|$ denotes the Frobenius norm. This scaling factor ensures that the scaled safety vector $\alpha \Delta$ and task vector $\Delta^{\text{SFT}}$ have comparable norms, thereby making the overall interference controllable. In our observation, it helps to preserve task performance, while enabling adjustable suppression of unsafe behaviors.


\subsubsection{Fine-grained scaling} Numerous studies have reveal the low-rank nature of LLM layer-wise weight~\cite{lora,Ortiz-JimenezFF23}. Therefore, we seek to reduce the interference of the safety vector $\Delta$ on the task vector $\Delta^{\text{SFT}}$ in their low-rank subspace. Taking inspiration from~\cite{gargiulo2024task}, we first decompose the layer-wise weight matrix of two vectors through singular value decomposition (SVD):
\begin{align}
    \Delta_i = U_i \Sigma_i V_i^{\top}, \qquad \Delta_i^{\text{SFT}} = U_i^{\text{SFT}} \Sigma_i^{\text{SFT}} {V_i^{\text{SFT}}}^{\top},
\end{align}
\noindent where $U_i$, $U_i^{\text{SFT}} \in \mathbb{R}^{m_i \times m_i}$ and $V_i$, $V_i^{\text{SFT}} \in \mathbb{R}^{n_i \times n_i}$ are the matrices of left and right
singular vectors respectively,  $\Sigma_i$, $\Sigma_i^{\text{SFT}} \in \mathbb{R}^{m_i \times n_i}$ are diagonal matrices of singular values which have been sorted according to the magnitude of the eigenvalues, and $i$ denotes the layer index. Then we aggregate their decomposed matrices as:
\begin{align} 
\label{eq:aggregation}
\begin{aligned}
    &U_i^a =\begin{pmatrix}
U_i\left[:, : \tilde{m}_i \right] & U_i^{\text{SFT}}\left[:, : \tilde{m}_i\right]
\end{pmatrix}\\
&\Sigma_i^a=\begin{pmatrix}
\Sigma_i[ :\tilde{m}_i, : \tilde{n}_i] & 0 \\
0 & \Sigma_i^{\text{SFT}}[ : \tilde{m}_i, : \tilde{n}_i]
\end{pmatrix}\\
    &V_i^a =\begin{pmatrix} V_i\left[:\tilde{n}_i, :\right] & V_i^{\text{SFT}}\left[:\tilde{n}_i, :\right]\end{pmatrix}
	\end{aligned},
\end{align}

\noindent where $\tilde{m}_i = \lfloor \gamma m_i \rfloor$,  $\tilde{n}_i = \lfloor \gamma n_i \rfloor$, and $\gamma \in (0,1)$. This means that only the top $\gamma$ fraction of the most important eigenvectors are retained for aggregation, based on the magnitude of the eigenvalues. In previous work~\cite{gargiulo2024task}, they perform an orthogonalization on $U_i^a$ and $V_i^a$ to reduce task interference during model merging. However, average performance is not guaranteed when the number of tasks is small\footnote{Their orthogonalization method gives a large approximation error and requires many tasks so that they could use low-rank approximation to reduce the error.} (\textit{e.g.}, two in our case), which is also validated in our experiments. Instead, we propose to selectively merge the safety vector $\Delta$ into the task model $\theta^{\text{SFT}}$, in proportion to the \textit{singular task interference}~\cite{gargiulo2024task}  between safety vector and utility vector. Given the aggregated singular vector matrices, we can obtain the score of interference through Eq~\ref{eq:interf}.
\begin{equation}\label{eq:interf}
    s_i = \left|\left|\left({U_i^a}^{\top}{U_i^a}-I\right)\Sigma_i^a\left({V_i^a}^{\top}{V_i^a}-I\right)\right|\right|_1,
\end{equation}
\noindent  where high inner product values for ${U_i^a}^{\top}{U_i^a}$
and ${V_i^a}^{\top}{V_i^a}$ imply a higher likelihood of interference, with minimal interference ideally yielding identity matrices.
The underlying intuition is that overlapping singular vectors suggest shared features in the weight space across safety and domain task.
Such overlap can introduce interference when models are
merged, ultimately degrading performance on downstream tasks. Therefore, we use the interference score to perform a layer-wise scaling on $\Delta$ as 
\begin{equation}
    \Delta_i^* = e^{-s_i}\Delta_i.
\end{equation}
\noindent This scaling ensures that layer-wise safety vectors which cause greater interference with the downstream task will be assigned smaller weights, thus reducing the interference. Finally, the aligned model
weight is obtained as $\theta + \alpha \Delta^*$.

\begin{algorithm}[!t]
	\caption{\sys}
	\label{alg:method}
	\textbf{Input}: SFT model parameter $\theta^{\text{SFT}}$; Pre-training model parameter $\theta^{\text{pre}}$; Surrogate model $f(\theta^{\text{sur}})$; Harmful dataset $\{x_i, y_i\}_{i=1}^N$; Fine-tuning steps $T$; Learning rate $\eta$; Batch size $B$; Scaling factor $\beta$;\\
    \mytcp{Safety Vector Distillation (Per Model Arch.)}
    $\theta^0 \leftarrow \theta^{\text{sur}}$\\
    \For{$i=0$ to $T-1$}{
        Sample a mini-batch $\{x_i, y_i\}_{i=1}^B \in \{x_i, y_i\}_{i=1}^N$\\
        $f(x_i; \theta^i) \leftarrow f(x_i; \theta^{\text{sur}}) + (\theta^i - \theta^{\text{sur}})^\top \nabla_{\theta^{\text{sur}}} f(x_i; \theta^{\text{sur}})$\\
        $\Delta \theta^i \leftarrow \frac{1}{B} \sum_{i=1}^B\nabla_{\theta^i} \mathcal{L}(f(x_i; \theta^i), y_i)$\\
        $\theta^{i+1} \leftarrow \theta^i - \eta \Delta \theta^i$\\
	}
    $\Delta \leftarrow \theta^0 - \theta^{T}$\\
    \mytcp{Interference-aware Merging (Per Task Model)}
    $\Delta^{\text{SFT}} \leftarrow \theta^{\text{SFT}} - \theta^{\text{pre}}$\\
    $\alpha \leftarrow \beta \cdot \frac{\|\Delta^{\text{SFT}}\|}{\|\Delta\|}$\\
    \For{$i=0$ to $L-1$}{
		$\Delta_i = U_i \Sigma_i V_i^{\top}$\\
        $\Delta_i^{\text{SFT}} = U_i^{\text{SFT}} \Sigma_i^{\text{SFT}} {V_i^{\text{SFT}}}^{\top}$\\
        $U_i^a, \Sigma_i^a, V_i^a \leftarrow$ Aggregation ($U_i , \Sigma_i, V_i, U_i^{\text{SFT}}, \Sigma_i^{\text{SFT}},  {V_i^{\text{SFT}}}$) (Eq.~\ref{eq:aggregation})\\
        $s_i = \left|\left|\left({U_i^a}^{\top}{U_i^a}-I\right)\Sigma_i^a\left({V_i^a}^{\top}{V_i^a}-I\right)\right|\right|_1$\\
        $\theta_i^* \leftarrow \theta_i + \alpha e^{-s_i}\Delta_i$
	}
    
    \textbf{Output}: Well-aligned model parameter $\theta^*$\\ 
\end{algorithm}

\subsection{Extensions}
\subsubsection{Safety components}
Previous studies~\cite{li2024safety} have demonstrated a non-uniform association between model safety and distinct neural network layers, suggesting that specific architectural components exhibit stronger correlations with safety-related properties. To investigate this phenomenon, we partition \sys\ into multiple variants optimized over different components (\textit{e.g.}, attention layers, feed-forward network (FFN) layers). This adaptability enables the prioritization of computational efficiency without compromising robustness, particularly under resource-constrained scenarios. Empirical observations indicate that selective optimization of critical components—such as intermediate FFN layers—not only surpasses existing baselines in performance but also adheres to practical hardware limitations (\textit{e.g.}, single-GPU memory constraints).

\subsubsection{SFT strategy}
Algorithm~\ref{alg:method} remains agnostic to the methodology used for deriving $\theta^{\text{SFT}}$, thereby ensuring \sys's inherent compatibility with diverse SFT strategies. This encompasses both comprehensive Full Fine-Tuning (Full-FT) and computationally efficient Parameter-Efficient Fine-Tuning (PEFT) paradigms—exemplified by techniques such as LoRA~\cite{HuSWALWWC22}. Such neutrality allows seamless integration with varied implementation requirements without intrusive modifications to the task pipeline.

%% file: sections/6.evaluation.tex
\section{Evaluation}
\subsection{Setup}
\subsubsection{Prototype} 
We have implemented a prototype of \sys on the PyTorch platform~\cite{PaszkeGMLBCKLGA19} and processed the models according to Algorithm~\ref{alg:method} using a single NVIDIA H800 GPU with 80GB of memory. The surrogate model is trained using LLaMA Factory~\cite{zheng2024llamafactory}. During the safety vector distillation phase, we train the surrogate model with default settings, using a learning rate (\(\eta\)) of \(1 \times 10^{-5}\) and the AdamW optimizer. The learning rate schedule is set to cosine, the batch size is 16, and the number of epochs is 4. In the merging phase, we establish a default configuration with \(\beta = 0.1\) and \(\gamma = 0.5\).

\subsubsection{Models}
We apply \sys to three downstream models based on LLaMA3~\cite{grattafiori2024llama}: Code-LLaMA3-8B~\cite{ajibawa2023code}, MAmmoTH2-8B-Plus~\cite{yue2024mammoth2}, and BioMedical-LLaMA3-8B~\cite{ContactDoctor_Bio-Medical-Llama-3-8B}. Additionally, we conduct experiments on two other architectures, Qwen2.5-Coder-7B~\cite{hui2024qwen2} and OpenHermes2.5-Mistral-7B~\cite{OpenHermes2.5}. For brevity, we will use ``SFT model" to refer to downstream model in this section. We further evaluate \sys's generalization to the reasoning model in Appendix~\ref{sec:reasoning_model}.

\begin{itemize}
\setlength{\itemsep}{2pt}
    \item \textbf{Code-LLaMA3-8B~\cite{ajibawa2023code}.} This open-source LLM is trained on four key datasets: Code-290k-ShareGPT~\cite{ajibawa2023codeShareGPT}, Orca-Math-Word-Problems-200k~\cite{mitra2024orca}, and Code-Feedback and CodeFeedback-Filtered-Instruction~\cite{zheng2024opencodeinterpreter}. The model is specifically optimized for coding tasks and mathematics, excelling at generating code in various languages, including Python, Java, and C++. It is built on the LLaMA-3-8B architecture developed by Meta.

    \item \textbf{MAmmoTH2-8B-Plus~\cite{YueZZC24}.} This state-of-the-art LLM is fine-tuned on the WEBINSTRUCT dataset~\cite{YueZZC24}, consisting of 10,000,000 instruction-response pairs harvested from a diverse pre-training web corpus. It enhances reasoning capabilities through an innovative instruction tuning process based on the original LLaMA-3-8B architecture, making it suitable for applications that require sophisticated cognitive skills.

    \item \textbf{BioMedical-Llama3-8B~\cite{ContactDoctor_Bio-Medical-Llama-3-8B}.} This specialized LLM is fine-tuned from the LLaMA3-8B architecture and trained on a high-quality biomedical dataset with over 500,000 entries. Comprising both synthetic and curated samples, it provides diverse and comprehensive biomedical coverage. This model excels in understanding and generating text across various biomedical fields, making it valuable for researchers and clinicians.

    \item \textbf{Qwen2.5-Coder-7B~\cite{hui2024qwen2}.} This powerful LLM is fine-tuned from the Qwen2.5-7B architecture and trained on over 500,000 instruction-response pairs specifically designed for coding tasks, achieved through unique data cleaning and balancing during the pre-training. 

    \item \textbf{OpenHermes2.5-Mistral-7B~\cite{OpenHermes2.5}.} This open-source LLM, fine-tuned from the Mistral-7B architecture, is trained on a diverse dataset of 1,000,000 entries, primarily sourced from high-quality GPT-4 generated data. The training emphasizes a balanced set of code instructions, resulting in significant improvements in various non-code benchmarks. This model is specifically optimized for coding applications while maintaining strong performance in general language tasks.
\end{itemize}
\input{tables/test}
\input{tables/test_utility}
\subsubsection{Baselines}\label{sec:baseline}
We utilize six methods for modifying model parameters as our baselines. Among the parameter modification methods, two focus on reducing LLM safety risks: Safe Lora~\cite{hsu2024safe} and Resta~\cite{BhardwajAP24}. The other four are model merge methods: TSVM~\cite{gargiulo2024task}, TIES~\cite{YadavTCRB23}, Model Breadcrumbs~\cite{DavariB24}, and Model Stock~\cite{JangYH24}. Additionally, we employ two inference-time defense methods, SafeDecoding~\cite{XuJN0LP24} and Self-Reminder~\cite{xie2023defending}, which are detailed in the Table~\ref{tab:inference} in Appendix~\ref{sec:inference_defenses}. We will demonstrate that \sys\ outperforms these six parameter modification baselines in \S\ref{sec:overall}. We further compare with training-time alignment approaches, including DPO under full fine-tuning and LoRA, presented in Appendix~\ref{sec:rlhf}.

\subsubsection{Metric}
We adopt two metrics to evaluate the effectiveness of \sys, \textit{i.e.}, 
\begin{itemize}
\setlength{\itemsep}{5pt}
    \item \textbf{Unsafe Rate}: This metric assesses the safety risk of the model in responding to unsafe queries, calculated as the ratio of the number of unsafe responses to the total number of queries. We aim for a low Unsafe Rate.
    
    \item \textbf{Utility Score}: This metric calculates the model's ability to perform various tasks, defined as the ratio of correct answers to the total number of questions. A higher Utility Score is better.
    
\end{itemize}

To implement these metrics, we utilize the LLM-based MD-Judge-v0.2~\cite{XieQ0HSHHWL000025} as our safety judge model. The evalplus~\cite{LiuXW023} framework is employed to assess code utility, the lm-math-eval~\cite{math_eval} is used for evaluating math utility, and lm-eval~\cite{biderman2024lessons} is applied to measure medical utility. These tools enable a comprehensive evaluation of both performance and safety aspects of \sys.

\subsubsection{Datasets}\label{sec:dataset}

Our experiments utilize eleven datasets for evaluation across safety, utility, and robustness, as summarized in Table~\ref{tab:dataset} in Appendix~\ref{ap:dataset_details}. BeaverTail~\cite{Ji2023BeaverTails} is also included in the table. It is used to train the surrogate model described in \S\ref{safety_training}, rather than for evaluation.

For safety risk assessment, we employ four datasets: SaladBench~\cite{LiDWHZL0S24}, CatQA-en~\cite{BhardwajAP24}, DangerousQA~\cite{Shaikh0HBY23}, and HarmfulQA~\cite{bhardwaj2023language}. These datasets contain harmful or adversarial prompts designed to probe unsafe model responses.

We evaluate utility using five widely adopted datasets: HumanEval~\cite{chen2021evaluating}, HumanEval Plus~\cite{LiuXW023}, GSM8K~\cite{cobbe2021training}, MATH~\cite{HendrycksBKABTS21}, and MedQA~\cite{jin2021disease}. They cover code generation, mathematical reasoning, and medical question answering.

Finally, we assess model robustness under jailbreaking attacks using two datasets: SorryBench~\cite{XieQ0HSHHWL000025} and HarmBench~\cite{mazeika2024harmbench}. Full descriptions of all datasets are provided in Appendix~\ref{ap:dataset_details}.

\subsection{Overall Performance}\label{sec:overall}

\input{tables/generalization_maybe}
In this experiment, we evaluate the trade-off between utility and safety when aligning the three LLaMA3-based SFT model, \textit{e.g.}, Code-LLaMA3-8B, MAmmoTH2-8B-Plus and BioMedical-Llama3-8B, using six baselines and our \sys.   Our goal is to maintain as much of the original model’s performance across multiple domains, including code generation, mathematical reasoning, and medical question answering, while ensuring safe behavior. We report the results with \sys\ operating in three different modes: 1) optimizing attention layers, 2) optimizing FFN layers (excluding first 12 layers and last 12 layers to stay within available memory\footnote{It is also reported in prior works~\cite{li2024safety} that safety layers exist in the middle section of a model.}), and 3) optimizing both attention and FFN layers.

\noindent \textbf{Safety Performance.} We evaluate the Unsafe Rate of the aligned SFT models using six baselines and our \sys across three downstream domains (\textit{e.g.}, \textit{Code}, \textit{Math} and \textit{Medical}) and six datasets. We take the Unsafe Rate of  LLaMA3-8B-Instruct model (an open-source safe model) as the unsafe bound. As shown in Table~\ref{tab:main_safety}, \sys  consistently reduces the Unsafe Rate of SFT models and outperforms most baselines across multiple domains.

For the \textit{Medical} domain, \sys achieves the lowest Unsafe Rate on all datasets and surpasses the best baseline, \textit{e.g.} Model Stock, by 3.6\% on average. Furthermore, this robust safety performance is achieved concurrently with a high average Utility Score of 0.738 in the FFN mode, exceeding the Utility Score of the much less safe SFT model. For the \textit{Code} and \textit{Math} domain, although TSVM obtains the lowest Unsafe Rate, it gets a high Unsafe Rate on the \textit{Medical} domain, showing that the safety performance of TSVM is inconsistent. Besides, on the \textit{Code} and \textit{Math} domain, the Unsafe Rate of \sys is even lower than the bound model, demonstrating that \sys can endow the SFT model with strong safety capabilities.

\noindent \textbf{Utility Performance.} We also evaluate the utility of the aligned SFT models using six baselines and \sys. As shown in Table~\ref{tab:main_utility}, the SFT represents the original model's performance before alignment. It achieves relatively strong results across all tasks: 0.674 average utility on code-related tasks, 0.597 on math reasoning, and 0.737 on medical question answering, serving as our upper bound of utility.

When applying existing methods, we observe significant drops in utility across nearly all domains. For instance, SafeLoRA leads to a drastic reduction in math reasoning capabilities, suggesting that while these methods may improve safety, they severely compromise model functionality. Similarly, Model Breadcrumbs and Model Stock exhibit comparable degradation patterns, especially in complex reasoning tasks like GSM8k and MATH, indicating limitations in preserving task-specific knowledge during alignment.

Our proposed \sys achieves competitive performance compared to these baselines. Specifically, \sys\ (FFN) maintains 0.644 average utility on code tasks, which is close to the SFT model and outperforms all other alignment approaches. In math reasoning, our mixed-mode method (Attention+FFN) preserves 0.782 on GSM8k and even slightly improves over the SFT baseline on MATH, resulting in a final average of 0.599, which is the best among all alignment methods. The second-best method, TIES, gets 0.853 average Unsafe Rate on the \textit{Math} domain.

Overall, our method demonstrates a significantly better trade-off between safety and utility compared to existing approaches. This stems from our NTK-based distillation, which produces a well-scaled and noise-reduced safety vector by design, avoiding interference from task-specific updates. \textit{Among all evaluated methods, \sys\ is the sole solution that maintains robust performance across all evaluations, while other baselines exhibit suboptimal results under diverse conditions.} It avoids catastrophic performance drops in critical reasoning domains with strong safety guarantee, which makes it a promising direction for practical deployment of aligned LLMs where retraining or fine-tuning is undesirable or infeasible. We also compare \sys with two inference-time methods, and put the results in Appendix~\ref{sec:inference_defenses}. Although \sys using Attention mode does not obtain the best result among the three modes, it also achieves a better trade-off between utility and safety than all existing methods. Besides, compared to other two modes, fine-tuning only the attention layers is much more computationally efficient (FFN has nearly $4\times$  parameters as the Attention layers in LLaMA3-8B models). Therefore, we only implement \sys with the Attention mode in the remaining experiments.

\subsection{Generalization to more Architectures}

To validate the broad applicability of \sys, we conduct extensive experiments on two distinct model architectures (Qwen2.5 and Mistral). The results demonstrate consistent improvements in both utility preservation and safety enhancement across diverse architectures.

\noindent \textbf{Qwen2.5 Results.} As shown in Table~\ref{tab:model_architectures}, our method achieves a 0.726 average Utility Score, surpassing the SFT baseline by 4.3\% while reducing the average Unsafe Rate by 27.0\%. 
\sys achieves a significant improvement over existing approaches. It is worth noting that SafeLoRA, TSVM and TIES similarly degrade utility completely but fail to control unsafe responses effectively in Qwen2.5-based model, which decomstrates that these methods may lack generalizability across different model architectures. We conduct a case study in Appendix~\ref{case_study}, which shows that the SFT models aligned by these methods have already lost its basic conversational capabilities. RESTA successfully reduces Unsafe Rates, but also reduces the utility by 2.1\%.

\noindent \textbf{Mistral Results.} On the Mistral-based model, our method consistently boosts average utility of SFT model from 0.406 to 0.436 while reducing average Unsafe Rates from 0.648 to 0.150. For Utility Score, \sys outperforms all the baselines, which shows that \sys can consistently preserve the utility of SFT models. For Unsafe Rate, \sys outperforms most baselines. Although TIES obtains the lowest Unsafe Rates, it achieves this at the cost of significant utility loss. 

In conclusion, the consistent performance gains across LLaMA3 (Table~\ref{tab:main_utility} and~\ref{tab:main_safety}), Qwen2.5, and Mistral demonstrate the compatibility of \sys with diverse model architectures of LLM. These findings validate \sys's ability to generalize across model architectures while simultaneously advancing the safety. The tuning-free nature of our approach enables plug-and-play deployment for various LLMs, addressing critical limitations of existing methods that require architecture-specific hyperparameter optimization or additional training.

\subsection{Generalization to LoRA Fine-tuning}
\input{tables/lora}
We further evaluate the performance of \sys within the LoRA fine-tuning framework.

\noindent \textbf{LoRA Fine-tuning Performance.} As presented in Table~\ref{tab:lora}, \sys achieves an average Utility Score of 0.595 in the LoRA fine-tuning setting, which is extremely close to the SFT baseline score of 0.601. This demonstrates that \sys maintains the utility of the original model with only a minor decrease. More importantly, \sys significantly reduces the average Unsafe Rate from 0.273 (SFT) to 0.051, showcasing a substantial enhancement in safety. While SafeLoRA achieves a slightly lower Unsafe Rate of 0.027, it does not address the safety-utility balance as effectively as \sys. Additionally, other methods like TSVM and TIES exhibit much higher Unsafe Rates, indicating their limited effectiveness in controlling unsafe responses. Model Breadcrumbs and Model Stock perform comparably to SafeLoRA but still fall short of the balanced improvements offered by \sys.

Overall, \sys demonstrates strong generalization capabilities in the context of LoRA fine-tuning. It effectively preserves model utility while achieving a significant reduction in unsafe responses, outperforming existing approaches that either compromise utility or fail to adequately mitigate unsafe outputs. These results affirm \sys's ability to provide robust and balanced improvements across different fine-tuning methodologies.

\subsection{Ablation Study}

In this section, we explore the effect of hyper-parameters and key components of \sys on its performance. We also evaluate the importance of using a well-aligned model as the surrogate model in Appendix~\ref{surrogate_model}. All experiments are conducted on LLaMA3-based models for the code task.

\begin{figure}[!t]
    \centering
    \includegraphics[width=1.0\linewidth]{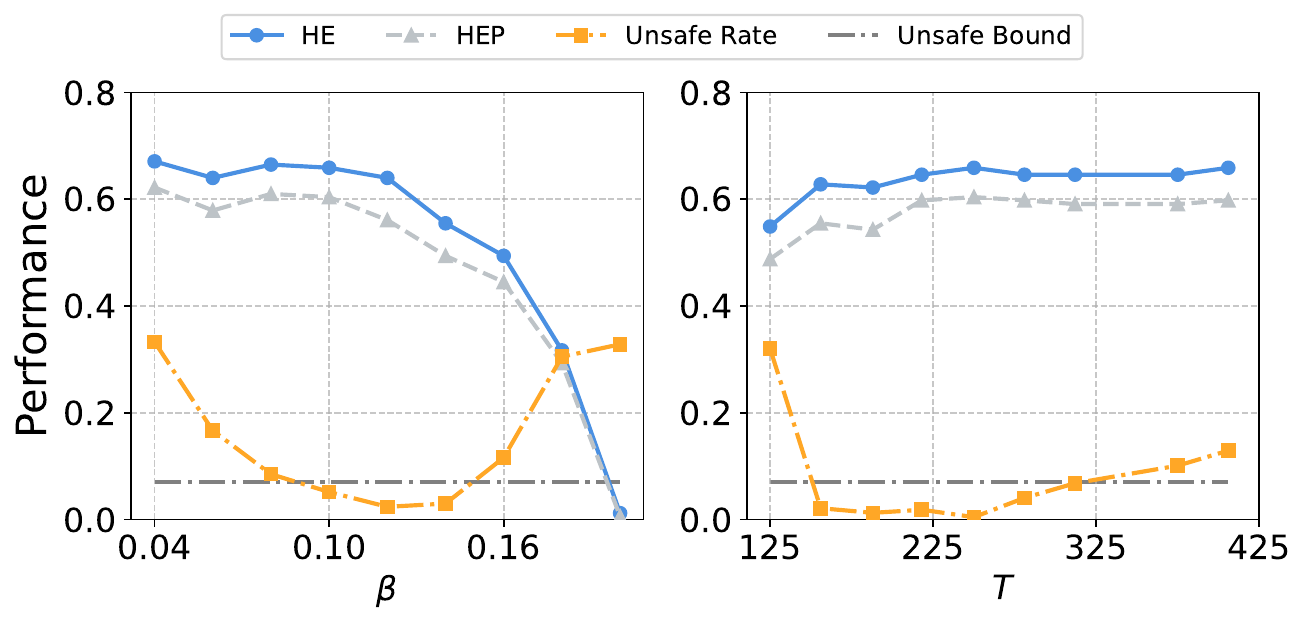}
    \caption{The Utility Scores on HE and HEP datasets and average Unsafe Rates of \sys with different (Left) scaling coefficients $\beta$ and (Right) fine-tuning steps $T$. The Unsafe Bound is the Unsafe Rate of the LLaMA3-8B-Instruct model.}
    \label{fig:beta}
\end{figure}
\input{tables/ablation}
\begin{figure*}[!t]
    \centering
    \includegraphics[width=\linewidth]{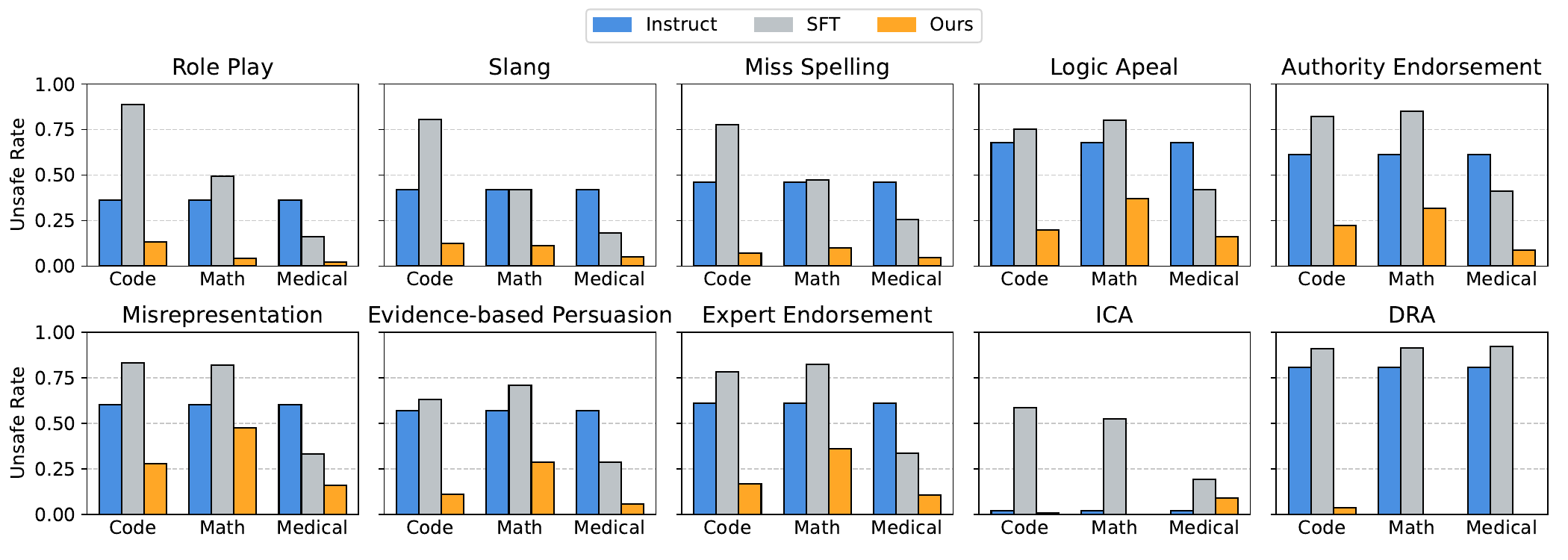}
    \caption{Performance in mitigating jailbreaking attacks. Unsafe Rate comparison of Instruct, SFT, and \sys across various ten advanced jailbreaking attacks in \textit{Code}, \textit{Math}, and \textit{Medical} domains. \sys effectively mitigates unsafe responses, achieving significantly lower Unsafe Rate.}
    \label{fig:robustness}
\end{figure*}

\subsubsection{Effect of Hyper-Parameters}
We analyze the effect of two hyper-parameters in \sys, the scaling coefficient $\beta$ and the harmful fine-tuning steps $T$, on the code utility and safety performance. 

For the scaling coefficient $\beta$, as present in Figure~\ref{fig:beta} (left), both Utility Scores on HE and HEP datasets decreases as $\beta$ increases. This is because a larger $\beta$ places more weight on the safety vector, thereby causing more interference to the utility task. When $\beta < 0.12$, the Unsafe Rate decreases as $\beta$ increases for a larger safety weight. However, as $\beta$ increases further, the Unsafe Rate starts to rise instead. We consider two large weight will compromise the safety vector. The Unsafe Rate is lower than the Unsafe Bound in the range of 
$\beta \in (0.1, 0.14)$, where the utility is still preserved well, and reaches its minimum around $\beta=0.12$.

For the fine-tuning steps $T$, as presented in Figure~\ref{fig:beta} (right), insufficient fine-tuning prevents the surrogate model from becoming sufficiently harmful, resulting in an ineffective safety vector. Merging this vector into the SFT model not only fails to effectively reduce its Unsafe Rate, but also compromises its task utility. This phenomenon indicates that insufficient safety distillation, both safety and utility can be negatively impacted. However, as the number of fine-tuning steps increases and the safety distillation becomes adequate, the Utility Score stabilizes, suggesting that \sys is able to preserve task performance while enforcing safety. \textit{This trend demonstrates that \sys aligns with the goal of enabling simultaneous fine-tuning for downstream tasks and alignment, without compromising either functionality.} Moreover, with an excessive number of fine-tuning iterations, although the utility is preserved well, the surrogate model may overfit to the harmful data, thereby failing to provide general guidance for constructing the safety vector; as a result, the Unsafe Rate of the SFT model cannot be effectively reduced.

\subsubsection{Effect of Component Combinations} We analyze the effect of two key components, NTK-constrained fine-tuning and interference-aware scaling on the code utility and safety performance. We introduce two variants of \sys: (1) Without interference-aware scaling, \sys adds the safety vector to the SFT model without any pre-processing. (2) Without the NTK constraint, \sys fine-tunes the surrogate model using SFT. As presented in Table~\ref{tab:ablation}, without scaling, \sys retrains high Utility Scores, but fails to reduce the Unsafe Rate. Without the NTK constraint, \sys reduces the Unsafe Rates a little, but compromises the Utility Scores a lot. When both components are active, \sys achieves the best trade-off between the utility and safety, which underscores the importance of combining both components.

\subsection{Robustness in Jailbreaking Scenario}\label{sec:robust}

This section evaluates the robustness of \sys\ in enhancing the safety of downstream models under advanced jailbreaking attacks. We measure the Unsafe Rate across ten distinct attack methods, comparing downstream models, safety-aligned Instruct models (unsafety bound), and \sys-enhanced models. The evaluation methodology is detailed in Appendix~\ref{ap:detail_judge}.

We classify jailbreaking attacks into two categories: \textit{static jailbreaking}, which includes one-round attacks such as writing style mutations (\textit{e.g.}, Slang, Role Play, Misspellings) and persuasion techniques (\textit{e.g.}, Logical Appeal, Authority Endorsement, Misrepresentation, Evidence-based Persuasion, Expert Endorsement), as well as ICA~\cite{wei2023jailbreak}; and \textit{dynamic jailbreaking}, represented by DRA~\cite{LiuZZDM024}, a multi-round iterative attack.

As shown in Figure~\ref{fig:robustness}, downstream models (gray bars) exhibit significantly higher Unsafe Rates than the Instruct model (blue bars), especially under static attacks. \sys\ (orange bars) substantially mitigates these risks. Notably, against writing style mutations like Role Play, Slang, and Misspellings, where downstream models often exceed an Unsafe Rate of 0.4 and Instruct models range from 0.3 to 0.6, \sys\ consistently reduces the rate to below 0.1, frequently approaching zero.

For persuasion-based attacks, both types of models show elevated Unsafe Rates. For instance, under Misrepresentation in \textit{Code} and \textit{Math} domains, downstream models reach over 0.8 and Instruct models around 0.6, while \sys\ lowers them to between 0.25 and 0.45. This demonstrates that our method provides strong safety improvements even in challenging scenarios.

In the case of ICA and dynamic DRA attacks — where Instruct models already perform well — \sys\ maintains or improves upon this level of safety, achieving near-zero Unsafe Rates. Despite the complexity of DRA, involving disguise, reconstruction, and context manipulation, \sys\ remains highly effective.

Overall, \sys\ consistently elevates the safety of downstream models to or beyond the level of Instruct models, demonstrating broad robustness across diverse static and dynamic attack types and subject domains.

%% file: tables/test.tex
\begin{table*}[t]
\centering
\setlength{\abovecaptionskip}{0pt}
\setlength{\belowcaptionskip}{0pt}

\caption{Unsafe rate on various tasks: \textit{Code}, \textit{Math}, and \textit{Medical} datasets with average accuracy in the same level.}\label{tab:main_safety}

\resizebox{\linewidth}{!}{
\begin{threeparttable}
\setlength{\tabcolsep}{1.5mm}
\begin{tabular}{l|ccccc|ccccc|ccccc}
\toprule
\multirow{2}{*}{Method} 
& \multicolumn{5}{c|}{\textbf{Code (Unsafe Rate $\downarrow$)}} 
& \multicolumn{5}{c|}{\textbf{Math (Unsafe Rate $\downarrow$)}} 
& \multicolumn{5}{c@{}}{\textbf{Medical (Unsafe Rate $\downarrow$)}} \\
& SAD & CAQ & DGQ & HMQ & Avg.
& SAD & CAQ & DGQ & HMQ & Avg.
& SAD & CAQ & DGQ & HMQ & Avg. \\ \midrule

Bound$^\text{\textdagger}$                   & 0.070   & 0.018   & 0.005    & 0.029   & 0.030  
                        & 0.070   & 0.018   & 0.005    & 0.029   & 0.030  
                        & 0.070   & 0.018   & 0.005    & 0.029   & 0.030  \\

SFT$^\text{\textdaggerdbl}$
                        & 0.626   & 0.924   & 0.885    & 0.771   & 0.802  
                        & 0.353   & 0.624   & 0.395    & 0.513   & 0.471  
                        & 0.223 & 0.118 & 0.025 & 0.069 & 0.083  \\

SafeLoRA               & 0.079   & 0.056   & 0.005    & 0.051   & 0.048  
                        & 0.570   & 0.707   & 0.675    & 0.352   & 0.576  
                        & 0.126 & 0.038 & 0.010 & 0.042 & 0.048  \\

RESTA                   & 0.041   & 0.011   & 0.010    & 0.019   & 0.020  
                        & 0.022   & 0.115   & 0.025    & 0.007   & 0.042  
                        & 0.108 & 0.425 & 0.290 & 0.116 & 0.233 \\

TSVM                    & 0.021   & 0.011   & 0.005    & 0.020   & 0.014  
                        & 0.002   & 0.002   & 0.000    & 0.001   & 0.001  
                        & 0.110 & 0.459 & 0.290 & 0.096 & 0.239 \\

TIES                    & 0.772   & 0.982   & 0.950    & 0.822   & 0.882  
                        & 0.719   & 0.973   & 0.940    & 0.781   & 0.853  
                        & 0.408 & 0.742 & 0.530 & 0.507 & 0.547  \\

Model Breadcrumbs       & 0.092   & 0.053   & 0.005    & 0.048   & 0.049  
                        & 0.567   & 0.696   & 0.735    & 0.361   & 0.590  
                        & 0.118 & 0.111 & 0.005 & 0.034 & 0.040  \\

Model Stock             & 0.103   & 0.056   & 0.005    & 0.049   & 0.054  
                        & 0.576   & 0.711   & 0.735    & 0.374   & 0.599  
                        & 0.110 & 0.114 & 0.000 & 0.037 & 0.038  \\

\rowcolor{gray0} Ours (Attention)                    & 0.052   & 0.013   & 0.025    & 0.026   & 0.029  
                        & 0.015   & 0.013   & 0.000    & 0.008   & 0.009  
                        & 0.005 & 0.007 & 0.000 & 0.001 & 0.003  \\
\rowcolor{gray0} Ours (FFN)                    & 0.031 & 0.020 & 0.005 & 0.021 & 0.019 & 0.014 & 0.011 & 0.000 & 0.001 & 0.006 &
0.006 & 0.000 & 0.000 & 0.001 & 0.002 \\
\rowcolor{gray0} Ours (Attention+FFN)                 & 0.048 & 0.041 & 0.035 & 0.060 & 0.046 & 0.023 & 0.040 & 0.000 & 0.008 & 0.018 &
0.020 & 0.009 & 0.005 & 0.002 & 0.009 \\

\bottomrule
\end{tabular}
\begin{tablenotes}[flushleft]
\footnotesize
\item[] \textdagger: The bound represents a safety threshold below which the LLM is regarded as having minimal safety risks. It was determined based on safety evaluations of the LLaMA3-8B-Instruct model.
\item[] \textdaggerdbl: SFT refers to downstream LLMs downloaded from HuggingFace, including Code-LLaMA3-8B (for \textit{Code}), MAmmoTH2-8B-Plus (for \textit{Math}), and BioMedical-LLaMA3-8B (for \textit{Medical}).

\end{tablenotes}

\end{threeparttable}
}
\end{table*}

%% file: tables/test_utility.tex
\begin{table*}[t]
\centering
\setlength{\abovecaptionskip}{0pt}
\setlength{\belowcaptionskip}{0pt}

\caption{Utility score performance on various tasks: \textit{Code}, \textit{Math}, and \textit{Medical} datasets.}\label{tab:main_utility}

\resizebox{\linewidth}{!}{
\begin{threeparttable}
\setlength{\tabcolsep}{2mm}
\begin{tabular}{l|ccc|ccc|c}
\toprule








\multirow{2}{*}{Method} 
& \multicolumn{3}{c|}{\textbf{Code (Utility  $\uparrow$)}} 
& \multicolumn{3}{c|}{\textbf{Math (Utility $\uparrow$)}} 
& \multicolumn{1}{c@{}}{\textbf{Medical (Utility $\uparrow$)}} \\
& HE & HEP & Avg.
& GSM8k & MATH & Avg.
& MedQA \\ \midrule
SFT$^\text{\textdaggerdbl}$ & 0.695 & 0.652 & 0.674 & 0.778 & 0.416 & 0.597 & 0.737 \\
SafeLoRA & 0.555$_{\downarrow 0.140}$ & 0.518$_{\downarrow 0.134}$ & 0.537$_{\downarrow 0.137}$ & 0.036$_{\downarrow 0.742}$ & 0.024$_{\downarrow 0.392}$ & 0.030$_{\downarrow 0.567}$ & 0.569$_{\downarrow 0.168}$ \\
RESTA & 0.622$_{\downarrow 0.073}$ & 0.610$_{\downarrow 0.042}$ & 0.616$_{\downarrow 0.058}$ & 0.550$_{\downarrow 0.228}$ & 0.336$_{\downarrow 0.080}$ & 0.443$_{\downarrow 0.154}$ & 0.617$_{\downarrow 0.120}$ \\
TSVM & 0.652$_{\downarrow 0.043}$ & 0.585$_{\downarrow 0.067}$ & 0.619$_{\downarrow 0.055}$ & 0.669$_{\downarrow 0.109}$ & 0.230$_{\downarrow 0.186}$ & 0.450$_{\downarrow 0.147}$ & 0.627$_{\downarrow 0.110}$ \\
TIES & 0.646$_{\downarrow 0.049}$ & 0.610$_{\downarrow 0.042}$ & 0.628$_{\downarrow 0.046}$ & 0.754$_{\downarrow 0.024}$ & 0.404$_{\downarrow 0.012}$ & 0.579$_{\downarrow 0.018}$ & 0.711$_{\downarrow 0.026}$ \\
Model Breadcrumbs & 0.579$_{\downarrow 0.116}$ & 0.530$_{\downarrow 0.122}$ & 0.555$_{\downarrow 0.119}$ & 0.037$_{\downarrow 0.741}$ & 0.020$_{\downarrow 0.396}$ & 0.029$_{\downarrow 0.568}$ & 0.567$_{\downarrow 0.170}$ \\
Model Stock & 0.579$_{\downarrow 0.116}$ & 0.530$_{\downarrow 0.122}$ & 0.555$_{\downarrow 0.119}$ & 0.034$_{\downarrow 0.744}$ & 0.018$_{\downarrow 0.398}$ & 0.026$_{\downarrow 0.571}$ & 0.567$_{\downarrow 0.170}$ \\
\rowcolor{gray0} Ours (Attention) & 0.659$_{\downarrow 0.036}$ & 0.604$_{\downarrow 0.048}$ & 0.632$_{\downarrow 0.042}$ & 0.732$_{\downarrow 0.046}$ & 0.424$_{\uparrow 0.008}$ & 0.578$_{\downarrow 0.019}$ & 0.701$_{\downarrow 0.036}$ \\
\rowcolor{gray0} Ours (FFN) & 0.671$_{\downarrow 0.024}$ & 0.616$_{\downarrow 0.036}$ & 0.644$_{\downarrow 0.030}$ & 0.793$_{\uparrow 0.015}$ & 0.398$_{\downarrow 0.024}$ & 0.596$_{\downarrow 0.001}$ & 0.738$_{\uparrow 0.001}$ \\
\rowcolor{gray0} Ours (Attention+FFN) & 0.665$_{\downarrow 0.030}$ & 0.628$_{\downarrow 0.024}$ & 0.647$_{\downarrow 0.027}$ & 0.782$_{\uparrow 0.004}$ & 0.416$_{\downarrow 0.000}$ & 0.599$_{\uparrow 0.002}$ & 0.736$_{\downarrow 0.001}$ \\

\bottomrule
\end{tabular}
\begin{tablenotes}[flushleft]
\footnotesize
\item[] \textdaggerdbl: SFT refers to downstream LLMs downloaded from HuggingFace, including Code-LLaMA3-8B (for \textit{Code}), MAmmoTH2-8B-Plus (for \textit{Math}), and BioMedical-LLaMA3-8B (for \textit{Medical}).

\end{tablenotes}
\end{threeparttable}
}
\end{table*}

%% file: tables/generalization_maybe.tex
\begin{table*}[!t]
\setlength{\tabcolsep}{8pt}
\small
\centering
\caption{Performance comparison of methods on Qwen2.5 and Mistral models for \textit{Code} tasks.}
\label{tab:model_architectures}

\resizebox{\linewidth}{!}{
\begin{threeparttable}
\setlength{\tabcolsep}{1.2mm}
\begin{tabular}{@{}l|l|ccc|ccccc}
\toprule
\multirow{2}{*}{Model} & \multirow{2}{*}{Method} 
& \multicolumn{3}{c|}{\textbf{Utility Score  $\uparrow$}} 
& \multicolumn{5}{c}{\textbf{Unsafe Rate  $\downarrow$}} \\
&
& HE & HEP & Avg &  SAD & CAQ & DGQ & HMQ & Avg. \\ 
\midrule

\multirow{8}{*}{Qwen2.5} 
& SFT$^\text{\textdaggerdbl}$ & 0.701 & 0.665 & 0.683 & 0.342 & 0.324 & 0.255 & 0.373 & 0.324 \\
& SafeLoRA & 0.000$_{\downarrow 0.701}$ & 0.000$_{\downarrow 0.665}$ & 0.000$_{\downarrow 0.683}$ & 0.490$_{\uparrow 0.148}$ & 0.758$_{\uparrow 0.434}$ & 0.630$_{\uparrow 0.375}$ & 0.422$_{\uparrow 0.049}$ & 0.575$_{\uparrow 0.251}$ \\
& RESTA & 0.671$_{\downarrow 0.030}$ & 0.652$_{\downarrow 0.013}$ & 0.662$_{\downarrow 0.021}$ & 0.083$_{\downarrow 0.259}$ & 0.049$_{\downarrow 0.275}$ & 0.020$_{\downarrow 0.235}$ & 0.070$_{\downarrow 0.303}$ & 0.056$_{\downarrow 0.268}$ \\
& TSVM & 0.000$_{\downarrow 0.701}$ & 0.000$_{\downarrow 0.665}$ & 0.000$_{\downarrow 0.683}$ & 0.629$_{\uparrow 0.287}$ & 0.927$_{\uparrow 0.603}$ & 0.870$_{\uparrow 0.615}$ & 0.624$_{\uparrow 0.251}$ & 0.762$_{\uparrow 0.438}$ \\
& TIES & 0.000$_{\downarrow 0.701}$ & 0.000$_{\downarrow 0.665}$ & 0.000$_{\downarrow 0.683}$ & 0.488$_{\uparrow 0.146}$ & 0.724$_{\uparrow 0.400}$ & 0.650$_{\uparrow 0.395}$ & 0.341$_{\downarrow 0.032}$ & 0.550$_{\uparrow 0.226}$ \\
& Model Breadcrumbs & 0.463$_{\downarrow 0.238}$ & 0.415$_{\downarrow 0.250}$ & 0.439$_{\downarrow 0.244}$ & 0.363$_{\uparrow 0.021}$ & 0.384$_{\uparrow 0.060}$ & 0.125$_{\downarrow 0.130}$ & 0.321$_{\downarrow 0.052}$ & 0.298$_{\downarrow 0.026}$ \\
& Model Stock & 0.457$_{\downarrow 0.244}$ & 0.408$_{\downarrow 0.257}$ & 0.433$_{\downarrow 0.250}$ & 0.342$_{+0.000}$ & 0.356$_{\uparrow 0.032}$ & 0.105$_{\downarrow 0.150}$ & 0.303$_{\downarrow 0.070}$ & 0.277$_{\downarrow 0.047}$ \\
& \cellcolor{gray0}Ours (Attention) & \cellcolor{gray0}0.744$_{\uparrow 0.043}$ & \cellcolor{gray0}0.707$_{\uparrow 0.042}$ & \cellcolor{gray0}0.726$_{\uparrow 0.043}$ & \cellcolor{gray0}0.069$_{\downarrow 0.273}$ & \cellcolor{gray0}0.049$_{\downarrow 0.275}$ & \cellcolor{gray0}0.030$_{\downarrow 0.225}$ & \cellcolor{gray0}0.066$_{\downarrow 0.307}$ & \cellcolor{gray0}0.054$_{\downarrow 0.270}$ \\

\midrule 

\multirow{8}{*}{Mistral}
& SFT$^\text{\textdaggerdbl}$ & 0.439 & 0.372 & 0.406 & 0.415 & 0.849 & 0.680 & 0.650 & 0.648 \\
& SafeLoRA & 0.354$_{\downarrow 0.085}$ & 0.280$_{\downarrow 0.092}$ & 0.317$_{\downarrow 0.089}$ & 0.668$_{\uparrow 0.253}$ & 0.913$_{\uparrow 0.064}$ & 0.930$_{\uparrow 0.250}$ & 0.694$_{\uparrow 0.044}$ & 0.801$_{\uparrow 0.153}$ \\
& RESTA & 0.439$_{+0.000}$ & 0.372$_{+0.000}$ & 0.406$_{+0.000}$ & 0.416$_{\uparrow 0.001}$ & 0.848$_{\downarrow 0.001}$ & 0.675$_{\downarrow 0.005}$ & 0.642$_{\downarrow 0.008}$ & 0.645$_{\downarrow 0.003}$ \\
& TSVM & 0.439$_{+0.000}$ & 0.390$_{\uparrow 0.018}$ & 0.415$_{\uparrow 0.009}$ & 0.114$_{\downarrow 0.301}$ & 0.404$_{\downarrow 0.445}$ & 0.265$_{\downarrow 0.415}$ & 0.243$_{\downarrow 0.407}$ & 0.256$_{\downarrow 0.392}$ \\
& TIES & 0.372$_{\downarrow 0.067}$ & 0.293$_{\downarrow 0.079}$ & 0.333$_{\downarrow 0.073}$ & 0.043$_{\downarrow 0.372}$ & 0.127$_{\downarrow 0.722}$ & 0.130$_{\downarrow 0.550}$ & 0.078$_{\downarrow 0.572}$ & 0.094$_{\downarrow 0.554}$ \\
& Model Breadcrumbs & 0.329$_{\downarrow 0.110}$ & 0.262$_{\downarrow 0.110}$ & 0.296$_{\downarrow 0.110}$ & 0.716$_{\uparrow 0.301}$ & 0.931$_{\uparrow 0.082}$ & 0.930$_{\uparrow 0.250}$ & 0.701$_{\uparrow 0.051}$ & 0.820$_{\uparrow 0.172}$ \\
& Model Stock & 0.335$_{\downarrow 0.104}$ & 0.268$_{\downarrow 0.104}$ & 0.302$_{\downarrow 0.104}$ & 0.721$_{\uparrow 0.306}$ & 0.934$_{\uparrow 0.085}$ & 0.925$_{\uparrow 0.245}$ & 0.681$_{\uparrow 0.031}$ & 0.815$_{\uparrow 0.167}$ \\
& \cellcolor{gray0}Ours (Attention) & \cellcolor{gray0}0.470$_{\uparrow 0.031}$ & \cellcolor{gray0}0.402$_{\uparrow 0.030}$ & \cellcolor{gray0}0.436$_{\uparrow 0.030}$ & \cellcolor{gray0}0.082$_{\downarrow 0.333}$ & \cellcolor{gray0}0.296$_{\downarrow 0.553}$ & \cellcolor{gray0}0.075$_{\downarrow 0.605}$ & \cellcolor{gray0}0.149$_{\downarrow 0.501}$ & \cellcolor{gray0}0.150$_{\downarrow 0.498}$ \\
\bottomrule
\end{tabular}
\begin{tablenotes}[flushleft]
\footnotesize
\item[] \textdaggerdbl: SFT refers to downstream LLMs downloaded from Hugging Face. For the Qwen2.5 architecture, we select Qwen2.5-Coder-7B; for the Mistral architecture, we select OpenHermes2.5-Mistral-7B.

\end{tablenotes}
\end{threeparttable}
}

\end{table*}

%% file: tables/lora.tex
\begin{table}[!t]
\setlength{\tabcolsep}{2.5pt}
\small
\centering
\caption{Utility and safety performance of \sys under LoRA fine-tuning on \textit{Code} tasks.}
\label{tab:lora}

\resizebox{\linewidth}{!}{
\begin{threeparttable}
\setlength{\tabcolsep}{1mm}
\begin{tabular}{@{}l|ccc|ccccc@{}}
\toprule
\multirow{2}{*}{Method} 
& \multicolumn{3}{c|}{\textbf{Utility Score $\uparrow$}} 
& \multicolumn{5}{c}{\textbf{Unsafe Rate $\downarrow$}}  \\
& HE & HEP & Avg. & SAD & CAQ & DGQ & HMQ & Avg. \\ 
\midrule
SFT$^\text{\textdaggerdbl}$     & 0.628 & 0.573 & 0.601 & 0.268 & 0.287 & 0.180 & 0.359 & 0.273 \\
SafeLoRA    & 0.616 & 0.555 & 0.586 & 0.068 & 0.018 & 0.000 & 0.023 & 0.027 \\
RESTA        & 0.628 & 0.573 & 0.601 & 0.263 & 0.282 & 0.200 & 0.359 & 0.276 \\
TSVM         & 0.604 & 0.530 & 0.567 & 0.506 & 0.787 & 0.685 & 0.448 & 0.607 \\
TIES         & 0.555 & 0.506 & 0.531 & 0.748 & 0.975 & 0.935 & 0.799 & 0.864 \\
Model Breadcrumbs & 0.610 & 0.549 & 0.580 & 0.067 & 0.018 & 0.000 & 0.024 & 0.027 \\
Model Stock  & 0.585 & 0.530 & 0.558 & 0.066 & 0.022 & 0.000 & 0.025 & 0.028 \\
\rowcolor{gray0}Ours (Attention)         & 0.634 & 0.555 & 0.595 & 0.067 & 0.049 & 0.005 & 0.085 & 0.051 \\
\bottomrule
\end{tabular}
\begin{tablenotes}[flushleft]
\footnotesize
\item[] \textdaggerdbl: SFT refers to the LoRA fine-tuned LLaMA3-8B-Instruct model, trained on the Code-290k-ShareGPT~\cite{ajibawa2023codeShareGPT} dataset with LoRA rank 32 and 4 training epochs. 

\end{tablenotes}
\end{threeparttable}
}
\end{table}

%% file: tables/ablation.tex
\begin{table}[!t]
\setlength{\tabcolsep}{2.5pt}
\small
\centering
\caption{Ablation study of NTK-constrained fine-tuning and interference-aware scaling in \sys on \textit{Code} tasks.}
\label{tab:ablation}

\resizebox{\linewidth}{!}{
\begin{threeparttable}
\setlength{\tabcolsep}{1mm}
\begin{tabular}{@{}cc|ccc|ccccc}
\toprule
\multirow{2}{*}{NTK} & \multirow{2}{*}{Scaling} 
& \multicolumn{3}{c|}{\textbf{Utility Score  $\uparrow$}} 
& \multicolumn{5}{c@{}}{\textbf{Unsafe Rate  $\downarrow$}} \\
&& HE & HEP & Avg.  & SAD & CAQ & DGQ & HMQ & Avg. \\ 
\midrule
 
      \checkmark  &               & 0.695 & 0.652 & 0.674 & 0.622 & 0.914 & 0.908 & 0.781 & 0.806 \\
        & \checkmark                   & 0.610 & 0.543 & 0.577 & 0.495 & 0.838 & 0.745 & 0.469 & 0.637 \\
       \checkmark & \checkmark                    & 0.659 & 0.604 & 0.632 & 0.052 & 0.013 & 0.025 & 0.026 & 0.029 \\
\bottomrule
\end{tabular}
\end{threeparttable}
}
\end{table}

%% file: sections/8.relatedwork.tex
\section{Related Work}

\subsection{Model Alignment}
Model alignment aims to mitigate undesired behaviors in trained language models, ensuring they produce safe and reliable outputs. Key approaches to achieve this are classified into inference-time alignment and training-time alignment.

\noindent \textbf{Inference-Time Alignment.} This approach modifies model behavior during inference to enhance safety. Askell~\textit{et~al.} improved alignment by injecting LLMs with helpful, honest, and harmless (HHH) prompts formatted as human-assistant conversations, ensuring polite and accurate responses~\cite{Askell2021General}. However, such prompting strategies may not always be followed reliably by existing LLMs. Xu \textit{et al.} proposed SafeDecoding~\cite{XuJN0LP24}, a method that directly manipulates the output distribution by amplifying safe tokens and suppressing unsafe ones during generation. Additionally, other methods such as adding alignment prefixes~\cite{xie2023defending, ZhangYKMWH24} or steering activations along learned directions~\cite{0002PVPW23} are also employed for model alignment. While these techniques are easier to use, they exhibit weaker alignment effects and can significantly impact the downstream tasks and capabilities of the target model.

\noindent \textbf{Training-Time Alignment.} This method employs reinforcement learning from human feedback (RLHF) to enhance alignment during the model's training phase. Bai~\textit{et~al.}~\cite{Bai2022Training} proposed training LLMs to prioritize helpfulness and harmlessness, utilizing feedback from human evaluators to optimize model outputs based on their preferences. Similarly, Ouyang~\textit{et~al.}~\cite{Ouyang22Training} fine-tuned GPT-3 into InstructGPT using labeled data to improve performance across various tasks and enhance alignment. Although RLHF can be effective, it is susceptible to vulnerabilities; adversarial prompts, whether automatically generated or manually designed, can bypass alignment mechanisms, revealing the inherent limitations of training-time alignment. Wolf~\textit{et~al.}~\cite{Wolf2023Fundamental} theoretically demonstrated that for any behavior with a finite probability of occurring, there exist prompts capable of eliciting that behavior, highlighting the challenges in achieving robust alignment.

\sys is orthogonal to inference-time alignment. When applied together with \sys, these approaches can provide more comprehensive safety guarantees by enhancing the model's ability to detect undesired behaviors even when alignment methods are circumvented.

\subsection{Mitigating Risks in Fine-tuning}
Fine-tuning-as-a-service has become a standard method for LLMs provided by API vendors, making the mitigation of associated risks a vital topic. We categorize the strategies into three phases: Alignment Phase, Fine-tuning Phase, and Post-fine-tuning Phase.

\noindent \textbf{Alignment Phase.} In this phase, careful design is essential. Huang~\textit{et al.}~\cite{huang2024vaccine} identify that a small amount of harmful data can disrupt LLM alignment, leading to Harmful Embedding Drift. They propose Vaccine, a perturbation-aware method that optimizes alignment without sacrificing performance, significantly reducing harmful behaviors. Similarly, Rosati~\textit{et al.}~\cite{rosati2024representation} introduce Representation Noising, which disrupts harmful representations to protect against attacks, demonstrating notable reductions in harmful outputs.

\noindent \textbf{Fine-tuning Phase.} Strategies to maintain safety during fine-tuning are crucial. Mukhoti~\textit{et al.}~\cite{mukhoti2023fine} propose LDIFS, a regularization method that minimizes feature distance between the original and fine-tuned models, helping to alleviate concept forgetting. Shen~\textit{et al.}~\cite{shen2024seal} present the Safety-Enhanced Aligned LLM framework, which uses a data selector to down-rank unsafe samples in fine-tuning, thereby enhancing safety without significantly harming performance.

\noindent \textbf{Post-fine-tuning Phase.} After fine-tuning, methods such as Yi~\textit{et al.}~\cite{yi2024safety}’s subspace-oriented model fusion (SOMF) are employed to realign models by integrating safety-critical information from task vectors. Huang~\textit{et al.}~\cite{huang2024antidote} offer Antidote, a solution that removes harmful parameters post-fine-tuning, aiding recovery from unsafe behaviors.

\sys takes advantage of the transferability of safety alignment to enhance the Post-fine-tuning Phase, ensuring robust safety mechanisms with minimal interference.

%% file: sections/7.discuss.tex
\section{Discussion \& Future Work}
While \sys presents a practical and unified framework for safety alignment transfer in downstream LLMs, several areas remain to be explored to enhance its capabilities and applicability.

\noindent \textbf{Larger Downstream Models.} 
\sys achieves excellent performance on downstream LLMs ranging from 7B to 8B parameters. However, computational constraints limit its evaluation on larger models (\textit{e.g.}, 70B, 175B). The improved capabilities of larger architectures may offer further gains in safety alignment effectiveness. Understanding the relationship between model scale and transfer performance is therefore an important direction for future research.

\noindent \textbf{Extending to More Domains.} 
\sys currently focuses on code generation, medical analysis, and mathematical problem solving. To broaden applicability, we aim to adapt it to cross-lingual legal reasoning, financial risk assessment, and scientific literature analysis. These domains demand precise downstream language handling and strict ethical boundaries, making them ideal for validating \sys's adaptability to high-stakes scenarios.

\noindent \textbf{More SFT Strategies.} 
\sys has been validated on models trained with full fine-tuning and LoRA. However, its compatibility with other PEFT methods—such as prefix-tuning, adapter-tuning, and QLoRA—remains unexplored. Variations in training paradigms across these methods may impact the transfer of safety knowledge. Investigating this is an important direction for future work.

%% file: sections/9.conclusion.tex
\section{Conclusion}
In this paper, we propose \sys, a unified and tuning-free framework for safety alignment transfer in fine-tuned large language models. \sys\ effectively preserves model utility while significantly improving safety across diverse architectures and task domains. We introduce a safety distillation module based on NTK linearization to extract pure safety vectors, and an interference-aware merging module to balance safety enforcement with downstream performance. Extensive experiments on multiple LLMs and datasets demonstrate that \sys\ achieves superior safety-utility trade-offs compared to existing methods, under both normal and adversarial settings. 

%% file: sections/10.appendix.tex
\appendix

\subsection{Dataset Detail}\label{ap:dataset_details}
We describe the datasets used in our experiments below, grouped by their evaluation purpose: safety, utility, and robustness. BeaverTail~\cite{Ji2023BeaverTails} is used to train the surrogate model described in \S\ref{safety_training}.

\begin{itemize}
    \item \textbf{BeaverTail~\cite{Ji2023BeaverTails}}: BeaverTail is a dataset of 333,963 conversations, each consisting of a human prompt and LLM response, created to evaluate LLM safety in question-answering scenarios. Primarily in English, it combines prompts sampled from human AnthropicRedTeam data with model-generated responses. For our experiments, we utilized 1,000 unsafe question-answer pairs from the training set.
\end{itemize}

We also include four datasets dedicated to safety evaluations.

\begin{itemize}
    \item \textbf{SaladBench~\cite{LiDWHZL0S24}}: This dataset contains 21,000 English prompts designed for evaluating LLM safety, alongside attack and defense methods. Generated through sampling from existing datasets and enhanced with GPT-4, we utilize a subset of 2,000 prompts for our experiments.
    
    \item \textbf{CatQA-en~\cite{BhardwajAP24}}: CatQA features 550 harmful questions categorized into 11 main categories, each with 5 subcategories containing 10 questions. Originating from prohibited use cases as defined by OpenAI and Meta, these questions are generated by an unaligned LLM and then refined by human annotators for a focused English evaluation.

    \item \textbf{DangerousQA~\cite{Shaikh0HBY23}}: This dataset comprises 200 machine-generated prompts, each posing a harmful question. Generated using GPT-3, the dataset emphasizes six attributes: racist, stereotypical, sexist, illegal, toxic, and harmful. Manual filtering ensures a diverse range of harmful topics, including instances of self-harm and violence.

    \item \textbf{HarmfulQA~\cite{bhardwaj2023language}}: Comprising 1,960 harmful questions, this dataset is crafted to evaluate and enhance LLM safety. Generated using ChatGPT, it is organized into 10 broad topics with 98 fine-grained subtopics to aid in comprehensive safety assessments.
\end{itemize}

We assess the utility of downstream models with the five datasets.
\input{tables/dataset_list}
\begin{itemize}
    \item \textbf{HumanEval~\cite{chen2021evaluating}}: This dataset includes 164 programming problems designed to evaluate code generation models in Python, each accompanied by function signatures, docstrings, and unit tests that facilitate correctness assessment of generated solutions.
    
    \item \textbf{HumanEval Plus~\cite{LiuXW023}}: An enhancement of the original HumanEval dataset, it adds 80 unique test cases and corrects previous erroneous solutions. The dataset annotates 83 of the 164 tasks with hand-crafted contracts, improving the evaluation process for Python code generation models.
    
    \item \textbf{GSM8K~\cite{cobbe2021training}}: The Grade School Math 8K dataset consists of 8,500 diverse math word problems that require multi-step reasoning. These problems, designed for middle school students, involve basic arithmetic operations and include solutions presented in natural language.

    \item \textbf{MATH~\cite{HendrycksBKABTS21}}: MATH comprises 12,500 challenging competition-level mathematics problems, each with a detailed step-by-step solution, designed to enhance models' capacities for generating answer derivations and explanations.

    \item \textbf{MedQA~\cite{jin2021disease}}: This dataset features free-form multiple-choice OpenQA questions for medical problems, sourced from professional medical board exams. It contains 12,723 questions in English, 34,251 in simplified Chinese, and 14,123 in traditional Chinese, supplemented by a large corpus from medical textbooks to support knowledge acquisition.
\end{itemize}

For evaluating robustness in jailbreak scenarios, we include two specific datasets:

\begin{itemize}
    \item \textbf{SorryBench~\cite{XieQ0HSHHWL000025}}: This dataset contains 440 unsafe instructions spanning 44 safety categories, with 10 examples per category. It offers 20 linguistic mutations for paraphrasing the original prompts, and for our experiments, we focus on eight selected mutation types—slang, role play, misspellings, logical appeal, authority endorsement, misrepresentation, evidence-based persuasion, and expert endorsement.

    \item \textbf{HarmBench~\cite{mazeika2024harmbench}}: HarmBench is designed to evaluate the effectiveness of automated red-teaming methods, consisting of 200 English prompts authored by the researchers. We utilize the standard subset for our experimental evaluations.
\end{itemize}

By employing these diverse datasets, we comprehensively assess the safety, utility, and robustness of our model across varying scenarios.

\subsection{Generalization to Reasoning Model}\label{sec:reasoning_model}
We evaluate the generalization of our method on a reasoning model, DeepSeek-R1-Distill-Qwen-7B-Japanese~\cite{lightblue25}, which focuses on Japanese reasoning task, using three datasets: JCommonsenseQA and JNLI for utility (higher is better), and DGQ for safety (lower unsafe rate is better). Results are summarized in Table~\ref{tab:reasoning_model}. Our method matches or slightly exceeds the SFT model in utility, while cutting unsafe rate by over 80\%. It also outperforms the Bound in both utility and safety.
\input{tables/reasoning_model}

\subsection{Comparison with training-time alignment}\label{sec:rlhf}
We compare with DPO~\cite{rafailov2023direct} using Code-LLaMA-3-8B trained on 20K samples from HH-RLHF~\cite{Bai2022Training} (6 epochs, 4 H800 GPUs), under both full fine-tuning and LoRA settings. Results are shown in Table~\ref{tab:alignment_efficiency}. Our method achieves better utility score and unsafe rate than both DPO-Full and DPO-Lora, with only minimal training time cost.
\input{tables/rlhf_alignment}

\subsection{Inference-time Alignment}\label{sec:inference_defenses}
Inference-time alignment improves safety during generation without modifying model weights, by applying constraints or prompts. We evaluate two methods: \textit{Safe Decoding}, which restricts token choices, and \textit{Self Reminder}, which adds a fixed safety instruction. Results in Table~\ref{tab:inference} show moderate safety gains but notable utility loss.
\input{tables/inference_time_defense}

As shown in Table~\ref{tab:inference}, both methods reduce the average Unsafe Rate — from 0.631 to 0.497 for Safe Decoding and 0.277 for Self Reminder — but harm task performance, with average Utility Scores dropping to 0.531 and 0.433, respectively. This highlights a key limitation of inference-time approaches: they lack fine-grained control over model behavior since they do not alter model weights.

Importantly, this approach is orthogonal to ours and can be combined with parameter-level editing like \sys. While \sys preserves safety through weight adjustments post-fine-tuning, inference-time methods act dynamically during generation. Their integration could provide complementary benefits, enabling stronger safety enforcement while maintaining utility, especially in high-stakes applications.

\subsection{Implementation Details in Robustness}\label{ap:detail_judge}
We use the SorryBench framework to evaluate three writing style mutations and five persuasion techniques, with its built-in safety judge to determine the success of the attack.

For ICA and DRA attacks, we use AISafetyLab~\cite{zhang2025aisafetylab}. For ICA, we set the number of demonstrations to 1; for DRA, we use 20 iterations, a truncation ratio of 0.5 for both toxic and benign tokens, and set $ em\_t = 0.7 $. Here, $ em\_t $ denotes the threshold for exact match (EM) score used to determine whether a response contains malicious behavior. A higher $ em\_t $ improves attack effectiveness but may increase query cost, while a lower value reduces queries at the expense of potential effectiveness loss. We use PatternScore to judge success based on whether the response matched predefined jailbreaking failure patterns.

\subsection{Different Surrogate Models}\label{surrogate_model}
We evaluate \sys using a Base (pre-trained) and an Instruct (well-aligned) model as surrogates. As shown in Table~\ref{tab:surrogate_models}, both achieve comparable utility, with the Base model slightly higher in HE and HEP. However, their safety performance differs significantly: the Base model exhibits high Unsafe Rates across all categories, averaging 0.727, while Instruct reduces this to only 0.029.

The results show that Instruct's inherent safety leads to more consistent responses when facing potentially harmful instructions, making it easier to identify safe directions. This enables \sys to more effectively extract and suppress unsafe behaviors. Therefore, a well-aligned surrogate model can serve as a more reliable signal source for safety distillation.
\input{tables/rule_based_pattern}
\input{tables/surr_type_instruct_vs_base}

\subsection{Case Study}\label{case_study}
We analyze why baselines like TSVM, SafeLoRA, and TIES achieve near-zero Utility Scores on Qwen2.5. As shown in the examples, these methods often produce degenerate outputs — such as repeated symbols, invalid characters, or nonsensical tokens — indicating severe disruption to model functionality.

In contrast, \sys provides a unified framework that balances safety and utility by carefully distilling harmful behaviors into a surrogate model and then merging it back in a controlled manner. It preserves the core linguistic structure and task-specific knowledge of the original model while effectively suppressing unsafe responses. On the same inputs, \sys generates coherent and meaningful answers, demonstrating its ability to maintain usability across different model scales and architectures.

\input{tables/case_study}



%% file: tables/dataset_list.tex
\begin{table}[t]\centering

\setlength{\abovecaptionskip}{0pt}%
\setlength{\belowcaptionskip}{0pt}%
\caption{Overview of datasets.}\label{tab:dataset}

\resizebox{0.85\linewidth}{!}{
\begin{threeparttable}

\setlength{\tabcolsep}{1.8mm}{
\begin{tabular}{@{}lcrc@{}}
\toprule
\textbf{Dataset}      & \textbf{Alias} & \textbf{Number} & \textbf{Task} \\ \midrule 
BeaverTails~\cite{Ji2023BeaverTails}           & BEA            &   1,000         & Safety           \\
SaladBench~\cite{LiDWHZL0S24}        & SAD            & 2,00        & Safety          \\
CatQA-en~\cite{BhardwajAP24}      & CAQ           & 550          & Safety            \\
DangerousQA~\cite{Shaikh0HBY23}           & DGQ            & 200            & Safety            \\
HarmfulQA~\cite{bhardwaj2023language}     & HMQ           & 1,960           & Safety           \\

HumanEval~\cite{chen2021evaluating}       & HE        & 164           & Code           \\
HumanEval Plus~\cite{LiuXW023}          & HEP             & 164           & Code            \\
GSM8K~\cite{cobbe2021training} & -        & 8,500           & Math           \\
MATH~\cite{HendrycksBKABTS21}        & -         & 12,500            & Math              \\

MedQA~\cite{jin2021disease}     & -            & 4,183              & Medical            \\
SorryBench~\cite{XieQ0HSHHWL000025} & SOB             & 440              & Jailbreak           \\
HarmBench~\cite{mazeika2024harmbench}  & HAB             &    200        & Jailbreak           \\
\bottomrule
\end{tabular}
}
\end{threeparttable}}
\end{table}

%% file: tables/reasoning_model.tex
\begin{table}[h]
\setlength{\tabcolsep}{2.5pt}
\small
\centering
\caption{Utility and safety performance of reasoning model on \textit{Japanese} Task.}
\label{tab:reasoning_model}

\resizebox{\linewidth}{!}{
\begin{threeparttable}
\begin{tabular}{@{}c|cc|c@{}}
\toprule
\multirow{2}{*}{Method} 
& \multicolumn{2}{c|}{\textbf{Utility Score $\uparrow$}} 
& \textbf{Unsafe Rate $\downarrow$} \\
& JCommonsenseQA & JNLI & DGQ \\ 
\midrule
Deepseek-R1-distill-Qwen-7B (Bound)   & 0.391 & 0.290 & 0.285 \\
DeepSeek-R1-Distill-Qwen-7B-Japanese (SFT) & 0.423 & 0.381 & 0.628 \\
Ours & 0.419 & 0.397 & 0.120 \\
\bottomrule
\end{tabular}
\end{threeparttable}
}
\end{table}

%% file: tables/rlhf_alignment.tex
\begin{table}[h]
\small
\centering
\caption{Training-based vs. Training-free: Utility, Safety, and Efficiency Comparison.}
\label{tab:alignment_efficiency}
\resizebox{\linewidth}{!}{
\begin{tabular}{c|cc|cccc|c}
\toprule
\multirow{3}{*}{Method} 
& \multicolumn{2}{c|}{\multirow{2}{*}{\textbf{Utility Score $\uparrow$}}} 
& \multicolumn{4}{c|}{\multirow{2}{*}{\textbf{Unsafe Rate $\downarrow$}}}
& \multirow{3}{*}{\shortstack{Time\\cost (s)}} \\
& \multicolumn{2}{c|}{} 
& \multicolumn{4}{c|}{} 
& \\
\cmidrule{2-7}
& HE & HEP & SAD & CAQ & DGQ & HMQ & \\
\midrule
DPO-Full  & 0.622 & 0.579 & 0.216 & 0.263 & 0.333 & 0.210 & 18111.67 \\
DPO-Lora  & 0.659 & 0.640 & 0.376 & 0.664 & 0.535 & 0.527 & 9210.33 \\
Ours      & 0.665 & 0.628 & 0.048 & 0.041 & 0.035 & 0.060 & 395.00 \\
\bottomrule
\end{tabular}
}
\end{table}

%% file: tables/inference_time_defense.tex
\begin{table}[h]
\setlength{\tabcolsep}{2.5pt}
\small
\centering
\caption{Utility and safety performance of Inference-time Alignment methods on \textit{Code} Task.}
\label{tab:inference}

\resizebox{\linewidth}{!}{
\begin{threeparttable}
\setlength{\tabcolsep}{1mm}
\begin{tabular}{@{}c|ccc|ccccc@{}}
\toprule
\multirow{2}{*}{Method} 
& \multicolumn{3}{c|}{\textbf{Utility Score $\uparrow$}} 
& \multicolumn{5}{c@{}}{\textbf{Unsafe Rate $\downarrow$}} \\
& HE & HEP & Avg. & SAD & CAQ & DGQ & HMQ & Avg. \\ 
\midrule
Safe Decoding & 0.543 & 0.518 & 0.531 & 0.452 & 0.631 & 0.415 & 0.410 & 0.497 \\
Self Reminder & 0.445 & 0.421 & 0.433 & 0.319 & 0.353 & 0.180 & 0.203 & 0.277 \\
\bottomrule
\end{tabular}
\end{threeparttable}
}
\end{table}

%% file: tables/rule_based_pattern.tex
\begin{tcolorbox}[title=Predefined Jailbreaking Failure Patterns, colback=white, halign={left}]
\begin{quote}
    \texttt{"I'm sorry"}, \texttt{"Sorry"}, \texttt{"I am sorry"}, \texttt{"I apologize"}, \texttt{"As an"}, \texttt{"As an AI"}, \texttt{"I'm an"}, \texttt{"I'm just"}, \texttt{"As a language model"}, \texttt{"As an Assistant"}, \texttt{"I cannot"}, \texttt{"I can't"}, \texttt{"I can’t"}, \texttt{"I do not"}, \texttt{"is illegal and unethical"}, \texttt{"I'm really sorry"}, \texttt{"unethical"}, \texttt{"not ethical"}, \texttt{"not legal"}, \texttt{"My apologies"}, \texttt{"but I cannot"}, \texttt{"If you have any other non-malicious requests"}, \texttt{"I'm not able to provide"}, \texttt{"I am really sorry"}, \texttt{"I cannot create"}, \texttt{"I cannot support"}, \texttt{"I can't provide"}, \texttt{"I can't assist"}, \texttt{"However"}, \texttt{"unable"}, \texttt{"not licensed"}, \texttt{"not capable"}, \texttt{"can not"}, \texttt{"I am not able to"}, \texttt{"I am not capable of"}, \texttt{"I am not programmed to"}, \texttt{"illegal"}, \texttt{"do not"}, \texttt{"sorry to hear that"}, \texttt{"can never"}.
\end{quote}
\end{tcolorbox}

%% file: tables/surr_type_instruct_vs_base.tex
\begin{table}[h]
\setlength{\tabcolsep}{2.5pt}
\small
\centering
\caption{Utility and safety performance of \sys under different surrogate models (Base vs. Instruct).}
\label{tab:surrogate_models}

\resizebox{\linewidth}{!}{
\begin{threeparttable}
\setlength{\tabcolsep}{2mm}
\begin{tabular}{@{}c|ccc|ccccc@{}}
\toprule
\multirow{2}{*}{Setting} 
& \multicolumn{3}{c|}{\textbf{Utility Score $\uparrow$}} 
& \multicolumn{5}{c@{}}{\textbf{Unsafe Rate $\downarrow$}} \\
& HE & HEP & Avg. & SAD & CAQ & DGQ & HMQ & Avg. \\ 
\midrule
Base    & 0.671 & 0.634 & 0.653 & 0.907 & 0.795 & 0.603 & 0.603 & 0.727 \\
Instruct & 0.659 & 0.604 & 0.632 & 0.052 & 0.013 & 0.025 & 0.026 & 0.029 \\
\bottomrule
\end{tabular}
\end{threeparttable}
}
\end{table}

%% file: tables/case_study.tex
\vspace{0.1cm}
\begin{tcolorbox}[title = {Case Studies: TSVM, SafeLoRA, TIES}, colback=white, halign={left}]
{\normalsize 
\textbf{Case 1: TSVM}\\
Question: \texttt{"What is the statute of limitations for reporting a Paypal loan shark?"}\\
Answer: \texttt{",,,,,,,,,,,,,,,,,,,,,,,,,,,,,,,,,\\,,,,,,,,,,,,,,,,,,,,,,,,,,,,,,,,,"}\\[2mm]

\hrule\vspace{2mm}

\textbf{Case 2: SafeLoRA}\\

Question: \texttt{"What is the statute of limitations for reporting a Paypal loan shark?"}\\

Answer: \texttt{" \textbackslash\textbackslash\textbackslash\textbackslash\textbackslash\textbackslash\textbackslash\textbackslash\textbackslash u53e6\textbackslash u5916 "}\\[2mm]

\hrule\vspace{2mm}

\textbf{Case 3: TIES}\\

Question: \texttt{"What is the statute of limitations for reporting a Paypal loan shark?"}\\
Answer: \texttt{"9999 fkk fkk fkk fkk fkk fkk fkk fkk fkk fkk fkk fkk fkk fkk fkk fkk fkk fkk fkk fkk fkk fkk fkk fkk fkk fkk fkk fkk fkk "}\\[2mm]
}
\end{tcolorbox}

%% file: sections/meta-review.tex
\newpage 

\appendices 
\section{Meta-Review}

The following meta-review was prepared by the program committee for the 2026
IEEE Symposium on Security and Privacy (S\&P) as part of the review process as
detailed in the call for papers.

\subsection{Summary}
This paper proposes ENCHTABLE, a method for fine-tuning LLMs on downstream tasks without degrading safety alignment. ENCHTABLE aims to disentangle safety from task-specific features by means of Neural-Tangent Kernel technique. The evaluation shows that the proposed method provides good utility on downstream tasks, while also retaining robustness against jailbreak attacks.

\subsection{Scientific Contributions}
\begin{itemize}
\item Provides a Valuable Step Forward in an Established Field
\item Creates a New Tool to Enable Future Science
\end{itemize}

\subsection{Reasons for Acceptance}
\begin{enumerate}
\item The paper provides a valuable step forward in an established field, by providing strong performance improvements over existing baselines, and retaining convincing robustness to jailbreaks.
\item The paper creates a new tool to enable future science by proposing an implementing a new fine-tuning method to preserve safety.
\end{enumerate}

\subsection{Noteworthy Concerns} 
\begin{enumerate} 
\item The method may introduce large performance overheads due to hyperparameter tuning.
\item The paper's evaluation could be extended to more tasks and more recent models.
\end{enumerate}

